\documentclass[twoside]{article}

\usepackage{PRIMEarxiv}

\usepackage[utf8]{inputenc} 
\usepackage[T1]{fontenc}    
\usepackage{hyperref}       
\usepackage{url}            
\usepackage{booktabs}       
\usepackage{amsfonts}       
\usepackage{nicefrac}       
\usepackage{microtype}      
\usepackage{lipsum}
\usepackage{fancyhdr}       
\usepackage{graphicx}       

\usepackage{algorithm}
\usepackage{algorithmic}

\usepackage{amssymb}
\usepackage{amsmath,amsfonts}

\title{Identifying Predictions That Influence the Future: Detecting Performative Concept Drift in Data Streams
}

\author{
  Brandon Gower-Winter, Georg Krempl, Sergey Dragomiretskiy, Tineke Jelsma, Arno Siebes \\
  Utrecht University \\
  8 Hiedelberglaan, Utrecht, 3584 CS, NL \\
  \texttt{b.gower-winter@uu.nl, g.m.krempl@uu.nl, sdragomiretsky@gmail.com,} \\ \texttt{tinekejelsma@gmail.com, a.p.j.m.siebes@uu.nl} \\
}

\begin{document}
\maketitle

\begin{abstract}
Concept Drift has been extensively studied within the context of Stream Learning. However, it is often assumed that the deployed model's predictions play no role in the concept drift the system experiences. Closer inspection reveals that this is not always the case. Automated trading might be prone to self-fulfilling feedback loops. Likewise, malicious entities might adapt to evade detectors in the adversarial setting resulting in a self-negating feedback loop that requires the deployed models to constantly retrain. Such settings where a model may induce concept drift are called performative. In this work, we investigate this phenomenon. 

Our contributions are as follows: First, we define performative drift within a stream learning setting and distinguish it from other causes of drift. 
We introduce a novel type of drift detection task, aimed at identifying potential performative concept drift in data streams. 
We propose a first such performative drift detection approach, called \textbf{C}hecker\textbf{B}oard \textbf{P}erformative \textbf{D}rift \textbf{D}etection (CB-PDD). We apply CB-PDD to both synthetic and semi-synthetic datasets that exhibit varying degrees of self-fulfilling feedback loops. Results are positive with CB-PDD showing high efficacy, low false detection rates, resilience to intrinsic drift, comparability to other drift detection techniques, and an ability to effectively detect performative drift in semi-synthetic datasets. Secondly, we highlight the role intrinsic (traditional) drift plays in obfuscating performative drift and discuss the implications of these findings as well as the limitations of CB-PDD.
\end{abstract}


\section{Introduction}

\textit{Performative Drift (PD)} is a specific type of Concept Drift that occurs when predictions, made by deployed models, affect the future distributions they predict on. Prevalent in many practical domains such as Recommender Systems \cite{MansouryAbdollahpouriPechenizkiy2020}, adversarial settings such as malware, fraud and spam detection, learning in such scenarios is called \textit{Performative Prediction} \cite{HardtMendler2023}. For the most part, research has focused on settings where the presence of PD is known \textit{a priori}. Consequently, a research gap has been formed whereby most work in the performative domain neglects the practical setting in which these performative models operate. These settings are categorised by high-volumes of heterogeneous, non-stationary and potentially transient data. Properties that appropriate themselves naturally to the domain of Stream and Online Learning \cite{GamaZliobaiteBifet2014}, \cite{LuLiuDong2018}.

The primary goal of this research is to join together these two topics (Performative Prediction and Stream Learning), so that we may take existing research from one to solve problems or limitations in the other. In particular, we address the problem of detecting PD in settings where its presence is not known beforehand. At first glance, using Concept Drift detectors seems appropriate, however they are limited in that they are unable to distinguish between drift types. This means that if a traditional drift detector in used in a performative setting, one will be unable to identify if any drift detections are triggered by PD or by some other type of intrinsic drift. We address this limitation in the state-of-the-art by introducing a first-of-its-kind performative drift detector called \textit{CheckerBoard Performative Drift Detection} (CB-PDD) and evaluate its efficacy under various experimental conditions. In short, this paper contributes the following:

\begin{enumerate}
    \item A definition of performative drift and illustration as to why it is different to intrinsic drift.
    \item A method for detecting performative drift.
    \item A first-of-its-kind data generator for evaluating performative drift detectors.
    \item An evaluation of CB-PDD across various experimental scenarios including a sensitivity analysis, comparison to other drift detection techniques, a robustness evaluation in settings with both performative and intrinsic drift, and a demonstration of the efficacy of CB-PDD on three datasets with imputed performative drift.
\end{enumerate}

\section{Background}

\subsection{\label{sect:DSOML}Data Streams and Online Machine Learning}

Read and {\v{Z}}liobait{\.e} \cite{ReadZliobaite2023} define a data-stream as a mode of access to a potentially infinite sequence of instances generated by some concept and delivered to a learning algorithm by an instance-delivery-process (IDP). This may then be formally written as follows:

Consider a random variable $X$ with distribution $P(X)$, a realisation (instance) of $X$ is then denoted as $x \sim P(X)$. A stochastic generating process for $X$ may then be written as $\{X_0, X_1, \ldots, X_n\}$ with realisations of $X_i$ during this process forming a sequence of instances: $x_0, x_1, \ldots, x_n$. 

In a supervised learning task, each instance $x_i$ is a tuple $x_i = (x_i, y_i)$ which represents a training sample. The generation of said training sample is modelled as $(x_i, y_i) \sim P(Xi_, Y_i)$ where $P(X_i, Y_i)$ is an abstraction of the concept (generating process) at some timestep $i$. Finally, a data-stream for some supervised learning task is defined as:

\begin{equation}\label{eq:SUPERVISEDSTREAM}
    ((x_0,y_0), (x_1,y_1), \ldots, (x_i, y_i),\ldots)
\end{equation}

Where each training sample $(x_i, y_i)$ is drawn from some concept $P(X_i, Y_i)$ at timestep $i$.

\subsection{\label{sect:CDRIFT}Intrinsic Concept Drift}

In the context of data-streams, the term Concept Drift is used to describe perturbations in the data a stream learning algorithm receives. Concept Drift comes in many forms \cite{LuLiuDong2018}, but they all fundamentally describe a change in the data-generating process ($P(X_i, Y_i)$). If the generating process changes suddenly, it is known as abrupt drift. If it changes slightly over time, it is known as incremental or gradual drift, and if these changes are repetitive (e.g. public transportation usage patterns over workweeks compared to weekends), the drift is known as reoccurring drift. Formally, Concept Drift may be described as:

\begin{equation}\label{eq:CONCEPTDRIFT}
    P(X_{i-1}, Y_{i-1}) \neq P(X_i, Y_i)  
\end{equation}

Where $P(X_i, Y_i)$ is the generating process of some data-stream at timestep $i$. When the generating process of a data-stream at two timesteps is not equivalent, Concept Drift has occurred. Concept Drift may also be modelled as:

\begin{equation}\label{eq:CONCEPTDRIFTMODEL}
    P(X_i, Y_i) \sim D(\mathcal{P})
\end{equation}

Where $\mathcal{P}$ is the set of all possible joint distributions of $X$ and $Y$ and $D$ is the distribution over them. Using Equation \ref{eq:CONCEPTDRIFT}, we may also define where Concept Drift occurred from. It may occur in the prior probabilities of the class labels $P(Y)$, in the conditional class labels $P(X \vert Y)$, and in the posterior probabilities of the classes as well $P(Y \vert X)$. Lastly, it is possible that the incoming distribution $P(X)$ also changes without affecting $P(Y \vert X)$ \cite{GamaZliobaiteBifet2014}.

\subsection{\label{sect:PDRIFT}Performative Prediction}

Perdomo et al. \cite{PerdomoZrnicMendler2020} introduced the concept of \textit{Performative Prediction}. Inspired by Economic Forecasting \cite{HardtMendler2023} and Strategic Classification \cite{HardtMegiddoPapadimitriou2016}, Performative Prediction is the act of training a learning algorithm whose predictions affect the distribution of future instances the algorithm predicts on.

The goal of performative prediction is to evaluate the \textit{performative risk} a model $\theta$ produces under loss function $\ell$.

\begin{equation}\label{eq:PERFORMATIVERISK}
    Risk(\theta, D) = \underset{Z \sim D(\theta)}{\mathbb{E}} \ell(Z;\theta)
\end{equation}

where $D$ maps model $\theta$ to the distribution of labelled instances it produces $Z = (X, Y)$. Generally, a learning algorithm will try to minimize performative risk. This is challenging for two reasons: The distribution map $D$ is dependent on model $\theta$ and assumed to be unknown in most cases.

To solve these two issues, a procedure called \textit{repeated risk minimization} (RRM) is introduced (See Equation \ref{eq:RRM}). RRM is simply an update rule that resembles a simple retraining process found in many Stream Learning tasks \cite{HeChenLi2011}.

\begin{equation}\label{eq:RRM}
    \theta_{i + 1} = arg\underset{\theta}{min}\underset{Z \sim D(\theta_i)}{\mathbb{E}} \ell(Z;\theta)
\end{equation}

erformative Prediction has two notions of optimality. The first is known as \textit{Performative Stability} (PS):

\begin{equation}\label{eq:PS}
    \theta_{PS} \in arg\underset{\theta}{min} \, Risk(\theta, D(\theta_{PS}))
\end{equation}

When a model is performatively stable, it appears optimal on the distribution it gives rise to. A performatively stable model does not need to be retrained and is considered to be in a state of equilibrium. The second notion of optimality is known as \textit{Performative Optimality} (PO):

\begin{equation}\label{eq:PO}
    \theta_{PO} \in arg\underset{\theta}{min} \, Risk(\theta, D(\theta))
\end{equation}

Unlike PS, a performatively optimal model must just be optimal on the distribution produced after its deployment. This is a moving target and may require RRM for repeated retraining. In general, performatively optimal models need not be performatively stable and performatively stable models need not be performatively optimal. However, it follows that $Risk(\theta_{PO}, D) \leq Risk(\theta_{PS}, D)$ for any performative optimum $\theta_{PO}$ and stable point $\theta_{PS}$ \cite{PerdomoZrnicMendler2020}.  

\subsection{\label{CONNECTINGDRIFTS}Concept Drift and Performative Prediction}

The similarities between Performative Prediction and Concept Drift are  apparent. Both fields are concerned with changing distributions over time as well as introducing retraining protocols to minimize the impact said distributions have on the performance of a deployed model. The key difference however, is that Performative Prediction is concerned with distribution changes that arise from the predictions made by a deployed model whereas the source of change is not necessarily relevant in Concept Drift. 

Perdomo et al. \cite{PerdomoZrnicMendler2020} described Concept Drift as a more general problem to solve than that of Performative Prediction. In this work, we explicitly consider Performative Prediction to be specialized (subset) task of Concept Drift research. More specifically, when the distribution changes of a data-stream are induced by a predictive model(s), we call it \textit{Performative Drift} (PD). Alternatively, when the distribution changes of a data-stream are induced by other sources, we call it \textit{Intrinsic Drift}.

Formally describing PD as a specialized process of Concept Drift is also intuitive. Recall Equation \ref{eq:CONCEPTDRIFTMODEL} which describes a Distribution map $D$ from which an instance generating process (concept) $P(X_i, Y_i)$ is drawn from: $P(X_i, Y_i) \sim D(\mathcal{P})$. In the performative setting future concepts are conditioned on the currently deployed model $\theta$:

\begin{equation} \label{ew:PERFORMATIVEDRIFT}
    P(X_{i+1}, Y_{i+1}) \sim D(\mathcal{P} \, \vert \, \theta_i)
\end{equation}

Note that with this formalization, the concept $P$ is equivalent to $Z$ used in the definition of Performative Risk (Equation \ref{eq:PERFORMATIVERISK}). It is also worth noting that Equation \ref{ew:PERFORMATIVEDRIFT} does not make assumptions about the presence of intrinsic drift. In fact, it is entirely possible that a system with performative drift may also have intrinsic concept drift.

\subsection{Feedback Loops and Performative Drift}

Alternatively, PD can be modelled using feedback loops \cite{TaoriHashimoto2023}. There are two primary types of feedback loops. The first is a self-fulfilling feedback loop:

\begin{equation}
    P(\bar{X} \, \vert \, x, y, \hat{y}, \bar{y})_i \, < \, P(\bar{X} \, \vert \, x, y, \hat{y}, \bar{y})_{i+1}
\end{equation}

and the second-type is a self-defeating feedback loop:

\begin{equation}
    P(\bar{X} \, \vert \, x, y, \hat{y}, \bar{y})_i \, > \, P(\bar{X} \, \vert \, x, y, \hat{y}, \bar{y})_{i+1}
\end{equation}

where $x \in X$ is an instance with label $y$, prediction $\hat{y}$ (from some model $\theta$) and \textit{target prediction} $\bar{y}$. In a self-fulfilling feedback loop, the probability of encountering instances similar to $x$ increases if $\hat{y} = \bar{y}$. Conversely, the probability of encountering instances similar to $x$ decreases in a self-defeating prophecy if $\hat{y} = \bar{y}$. Note that feedback loops are not induced by an instance's label, but by its target prediction which may be triggered by either a correct classification ($y = \bar{y})$ or misclassification ($y \neq \bar{y}$).

The last component of the equation is $\bar{X}$ which is defined as the set of instances that are at least $\epsilon$ close to instance $x$ using some distance metric (e.g. Euclidean): 

\begin{equation}
    \bar{X} = \{\bar{x} \, \in \, X \; \verb|where |dist(x, \bar{x}) \, \leq \, \epsilon\}
\end{equation}

This distinction is important because the entity that produced $x$, and received feedback from the predictive model, may be capable of producing a variety of future instances that may not look exactly like instance $x$. As we show later, this dynamic can make it easier or harder to detect PD.

\section{CheckerBoard Detection}

\begin{figure}[t]
\centering
\includegraphics[width=0.85\columnwidth]{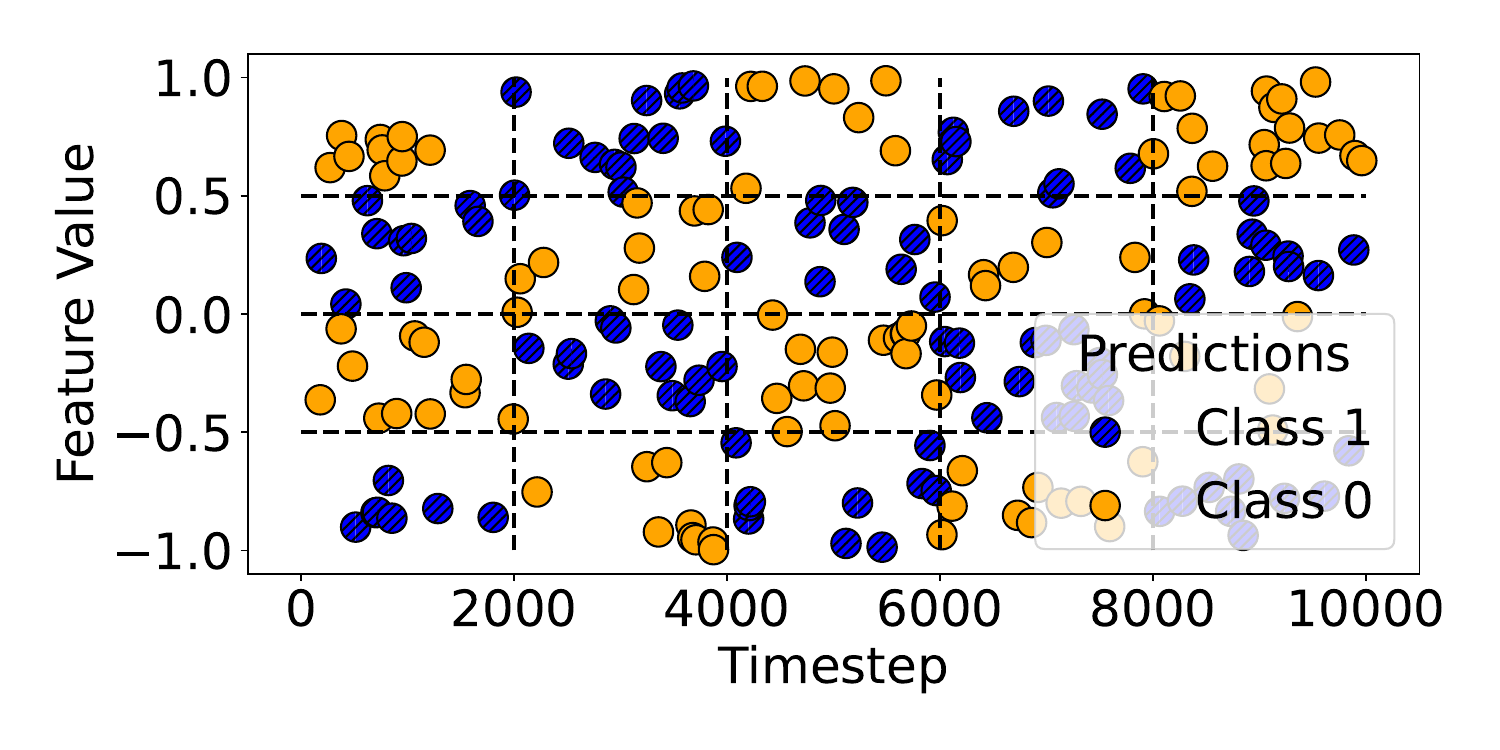}
\caption{Example of the Checkboard Pattern produced by the Checkerboard Detector for an arbitrary feature value in a Binary Classification task. In this Figure, the CheckerBoard Detector is parameterized with $f=0.5$ and $\tau = 2000$.}
\label{fig:CheckerBoardDetection}
\end{figure}

\textit{CheckerBoard Performative Drift Detection} (CB-PDD) is a three-stage algorithm which takes inspiration from controlled experiments and A/B testing \cite{KohaviLongbothamSommerfield2009}. In the first stage, incoming instances are classified according to a parameterized CheckerBoard pattern (Figure \ref{fig:CheckerBoardDetection}) instead of a predictive model. Given some data-stream of length $T$, a user-defined parameter $f$ is used to split a feature into groups. These groups are equivalent to the possible labels in a classification task (e.g. in a binary classification task, each feature is split according to $f$ and assigned either a label of $1$ or $0$). When an instance is received, the CheckerBoard assigns the instance as having a predicted value equivalent to the group it belongs to. After a predefined length of time $\tau$ (called trial length), the labels associated with each group are altered (e.g. in a binary classification task, a group that had incoming instances predicted as class $1$, will now be predicted as class $0$). It is this alternating of a group's assigned label that enables CB-PDD to perform A/B testing on the incoming instances. By forcing prediction according to the CheckerBoard, CB-PDD can check whether regions in the feature space are subject to PD. If a self-fulfilling feedback loop is present, increases in instance density should be observed in regions where the checkerboard's predictions ($\hat{y}$) match the target prediction ($\bar{y}$). In the case of a self-defeating feedback loop, the instance density will decrease instead.

After $T$ instances, stage two begins whereby the relative density changes are calculated for each trial period ($\tau$). This is done by creating two windows (of size $w$) which contain the first and last $w$ instances from a trial period. The relative density change is then calculated per class using these windows. The density changes are then stored for statistical testing. Two density values are calculated per trial per class label, The first value is the density changes that occurred when instances were predicted correctly (Group A), while the second contains the density changes where instances were predicted incorrectly (Group B). In the third stage, a statistical test (e.g. Mann-Whitney U test) is performed on Groups A and B using a user-defined confidence parameter $\alpha$. If the p-value returned from the test is less than $\alpha$, a detection event is triggered. Otherwise, no event is triggered as no performative drift was detected in the system. Note that in this work, CB-PDD is not presented as an online algorithm, this is because we only conduct the one statistical test after $T$ instances have been evaluated. The algorithm can be altered to operate in an online fashion with sliding windows and multiple statistical evaluations (such as in KSWIN \cite{RaabHeusingerSchleif2020}). This was deemed out of scope for this research as it introduces additional parameters to manage the sliding window and introduces the problem of repeated hypothesis testing.

\section{Synthetic Data Generator}

There is no consensus on how to model PD within a data-stream and, to the best of our knowledge, there exists no datasets with detected PD or data generators that support PD. In Perdomo et al. \cite{PerdomoZrnicMendler2020}, they use the deployed model's parameters $\theta$ to model performative drift over time. This approach is limited in that it assumes that the user has access to $\theta$, and is not compatible with approaches that impose alternative classification regimes such as CB-PDD.

To address this, we developed a model agnostic data generator for performative settings. Inspired by \textit{Random Radial Basis Function (RandomRBF)} generators \cite{MontielReadBifet2018}, our generator is initialized by specifying $C$ centroids within a feature space. Each centroid $c_i$ is given a Gaussian distribution, a label $y_i$ and a weight $w_i$. When an instance is requested, roulette wheel selection is performed \cite{LipowskiLipowska2012} using the weights. The Gaussian of the selected centroid is then sampled generating an instance $x$ which is assigned the corresponding label $y_i$.

In order to simulate PD, a user is required to supply a predictor to the data generator and the intended feedback loop behaviour for each class (i.e. The target prediction $\bar{y}$ for each class and whether a self-defeating or self-fulfilling feedback loop should be used). When an instance is generated, the predictor is queried and its prediction $\hat{y}$ is stored. If $\hat{y} = \bar{y}$, the weight $w_i$ of centroid $c_i$ is modified. In the case of a self-fulling feedback loop, $w_i = w_i + \sigma$. In the case of a self-defeating feedback loop, $w_i = w_i - \sigma$ where $\sigma$ is a user-defined parameter that determines the performative drift strength of the system.

\section{Experiments and Results}
Using the data generator introduced in the previous Section, we evaluate CB-PDD in several experimental settings. Unless stated otherwise, $n=50$ repetitions are conducted per parameter set evaluated and the \textit{detection rate} is reported. $T=100 000$ instances are generated for each repetition, and a confidence value ($\alpha = 0.01$) is used for all statistical tests to determine if PD is present within a data-stream. The statistical test for this process is the Mann-Whitney U test. By default $f=1.0$, $\tau = 1000$ and $w=100$

For simplicity, all of these scenarios are binary classification tasks with a single feature value in which both class labels exhibit self-fulfilling feedback loops. The data generator is initialized with $100$ centroids per class label (200 total) that are placed equidistantly across a feature space with range $[-1, 1]$. Each centroid is initialized with a random weight $\sim U(0,1)$ allowing for a wide range of starting distributions to be evaluated.

Additional experiments including self-defeating feedback loops, and scenarios where only one class label is performative are included in the Appendix of the paper along with a detailed parameter list for each set of experiments.

\subsection{Exploration of Trial Length ($\tau$)}

\begin{figure}[t]
\centering
\includegraphics[width=0.85\columnwidth]{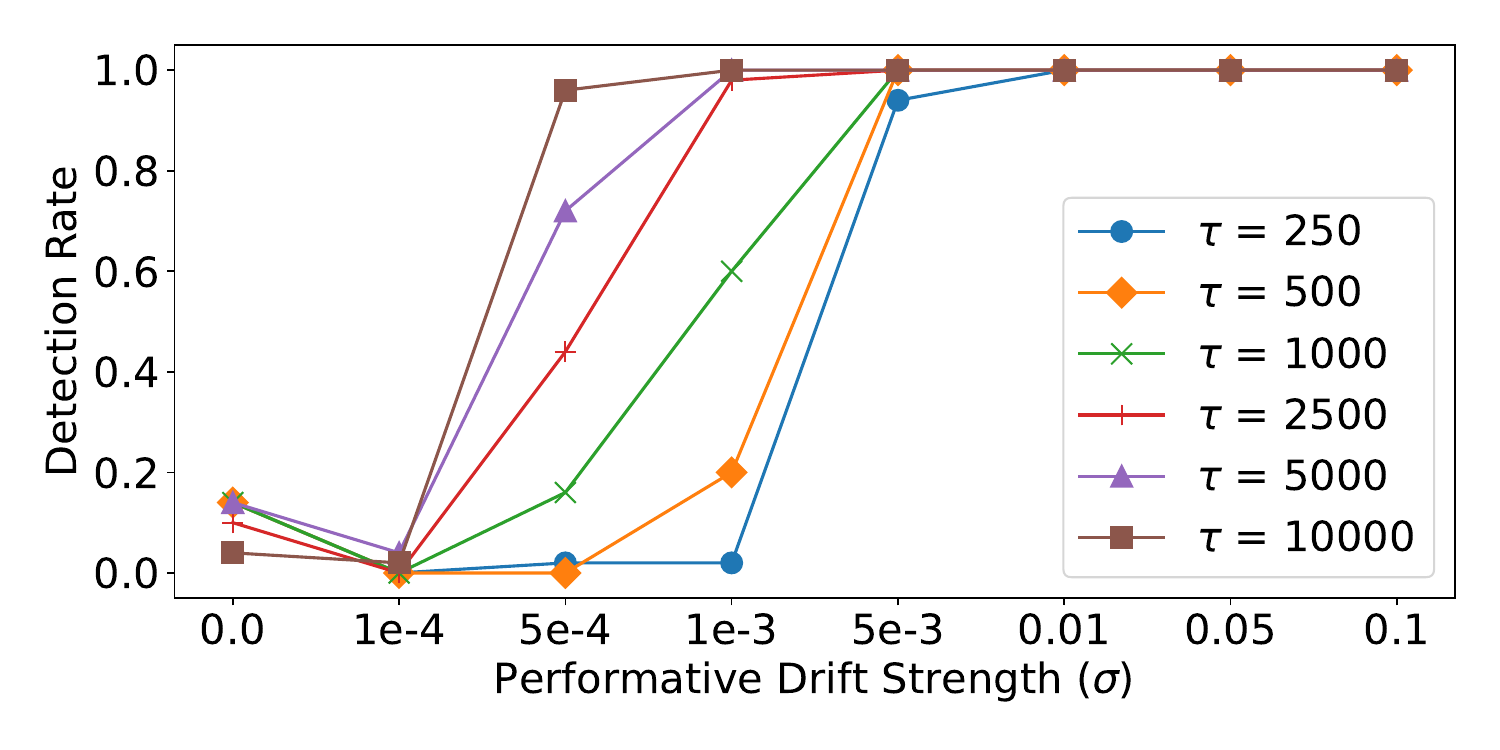}
\caption{shows the effect trial length ($\tau$) has on the detection rate of CB-PDD across various PD strengths ($\sigma$). The main finding is that as $\tau$ increases, so does the  detection rate.}
\label{fig:TAUEXPLORE}
\end{figure}

The first set of experiments we conducted explored the effect trial length ($\tau$) has on the detection rate of CB-PDD. Intuitively, as $\tau$ increases, so should the detection rate across all PD strength $\sigma$ values investigated. This is due to the increased time in which a classification regime is imposed by the CheckerBoard, allotting more time for a feedback loop to develop. This theory is corroborated by our results (Figure \ref{fig:TAUEXPLORE}), where as $\tau$ increases, so does the detection rate, regardless of $\sigma$. Conversely, our results show that if $\tau$ and $\sigma$ are small, CB-PDD is unlikely to detect any performative drift. This is reflected clearly in the $\tau=2500$ scenario where when $\sigma \leq 0.001$, we see a decline in the detection rate.. 

Lastly, we want to note the detection rate when $\sigma=0.0$ (i.e no performative drift). Across all of the experiments conducted in this work, the false detection rate (i.e. when no performative drift is present but a drift detection event is still triggered) for CB-PDD remains between $0\%$ and $20\%$. Furthermore, these false detections are often only raised by one of the classes and not both. These results are promising as false detections are a common issue in Concept Drift detection research which CB-PDD seems robust to.

\subsection{Exploration of Feature Split ($f$)}

\begin{figure}[t]
\centering
\includegraphics[width=0.85\columnwidth]{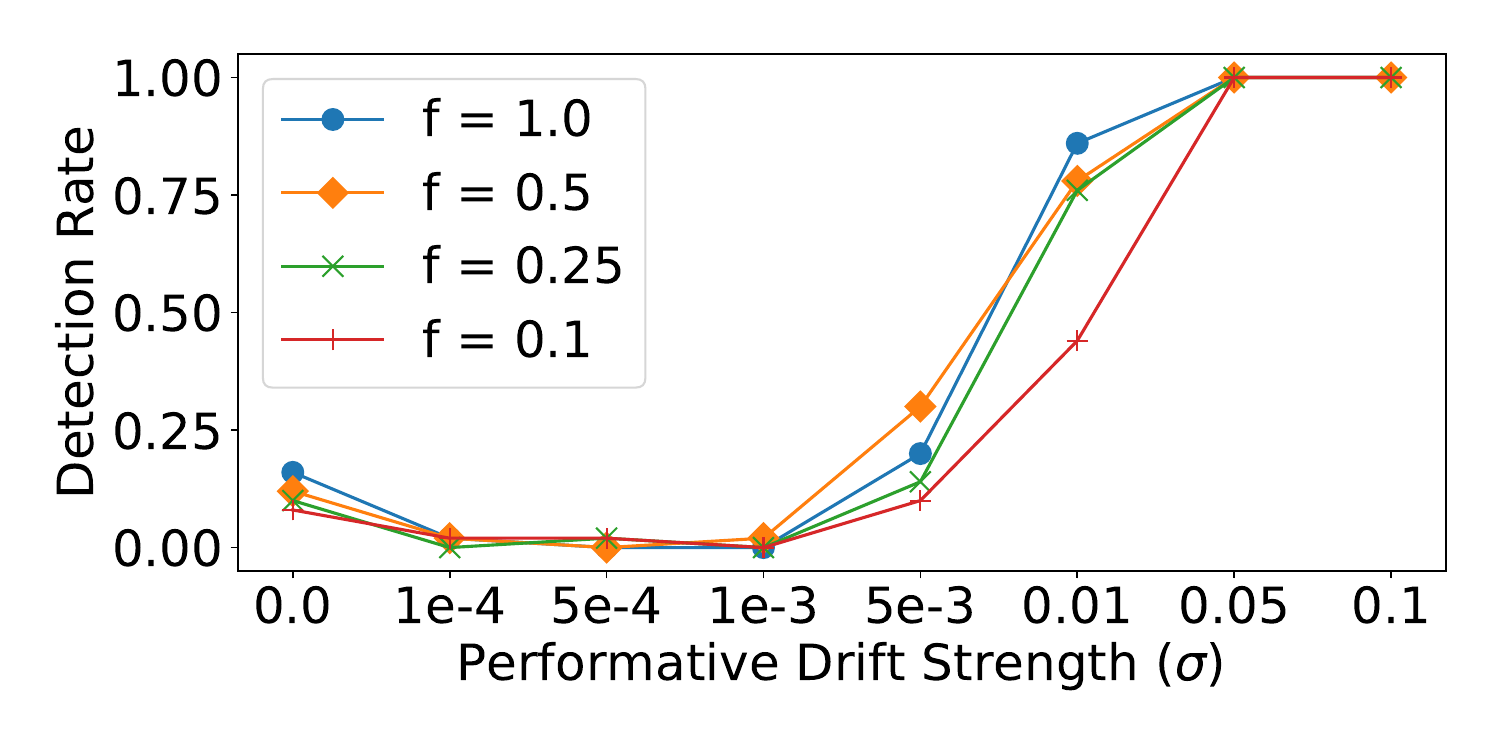}
\caption{shows the effect $f$ has on the detection rate of CB-PDD across various PD strengths ($\sigma$) in a low $\epsilon$ setting.}
\label{fig:F100EXPLORE}
\end{figure}

The second set of experiments explored the feature split $f$ parameter. At first glace, it is not clear how $f$ affects the detection rate. $f$ is required to partition a feature space into groups used by CB-PDD. Technically, if the range of a feature is $[-1,1]$, any $f \in (0,1.0]$ is valid, and any $f \in (0, 0.5]$ is valid for a feature with range $[0, 1]$. Figure \ref{fig:F100EXPLORE} illustrates this point where, for all $f$ values investigated, the detection rate remains consistent across all $\sigma$ investigated.

\begin{figure}[t]
\centering
\includegraphics[width=0.85\columnwidth]{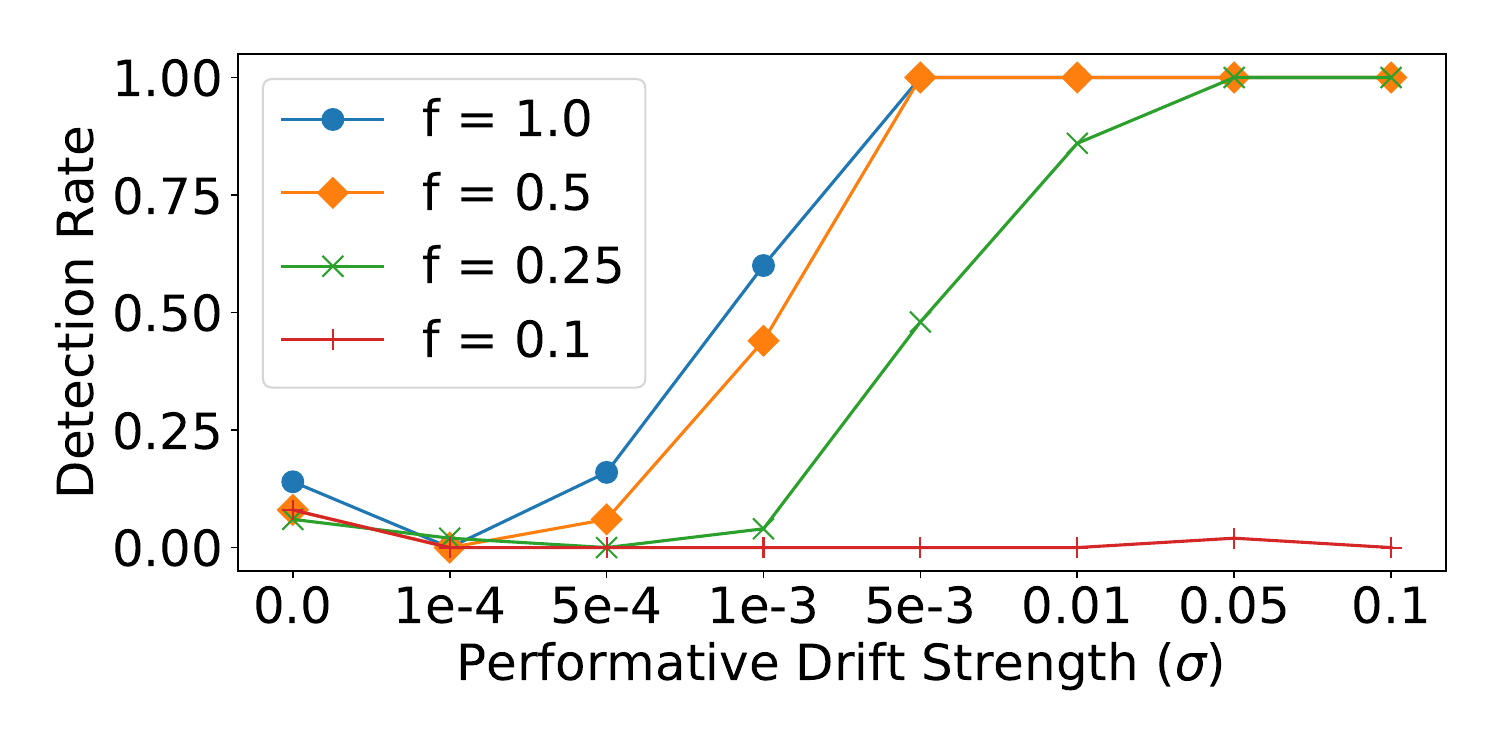}
\caption{shows the effect $f$ has on the detection rate of CB-PDD across various PD strengths ($\sigma$) in a high $\epsilon$ setting.}
\label{fig:F10EXPLORE}
\end{figure}

These results then seem to imply that $f$ is not an important parameter, but that is not the case. Recall that we introduced the concept of $\epsilon$ to describe the range of future instances that might be generated by inducing a feedback loop in some instance $x$. If $\epsilon$ is high, future instances generated by a feedback loop induced with $x$ will look different to $x$. The effect this has on $f$ is interesting, if $\epsilon$ is high and $f$ is low, a feedback loop induced by CB-PDD on instance $x$ may generate future instances that are so dissimilar to $x$ that they are assigned to a different group by CB-PDD, and thus assigned a different prediction. This has a negative effect on detection rate as inconsistent prediction assignment may reduce the momentum of a feedback loop or destroy it completely. We conducted additional experiments where we reduced the number of centroids to $10$ per class label (instead of $100$) and increased the spread of the Gaussians assigned to each centroid (increasing $\epsilon$). Results are reported in Figure \ref{fig:F10EXPLORE} and show this phenomena. As $f$ decreases, the detection rate across all but the strongest PD strength settings $\sigma = \{0.05, 0.1\}$ decreases too. The takeaway from these findings are that, in general, $f$ should be kept as high as possible to reduce the effect an unknown $\epsilon$ might have on the detection capabilities of CB-PDD.

\subsection{CB-PDD with Classifiers}

Traditional drift detectors typically work with a predictive model to detect drift. However, CB-PDD requires that some portion of the incoming instances be classified in accordance with the CheckerBoard predictor. Given this, we investigated how well CB-PDD performs when paired with a predictive model. To achieve this, we ran experiments where instances are randomly assigned to either the CheckerBoard predictor or the deployed predictive model in accordance with a $mix \in [0,1]$ parameter which describes the likelihood of an instance being given to the deployed predictor. 

We investigated two predictive models: a simple Random Classifier (RC) which randomly predicts which class ($1$ or $0$) an incoming instance belongs too, and a Threshold Classifier (TC) which predicts an instance's label in accordance with the following decision rule: $1 \verb| if | x > 0  \verb| else | 0$. The RC acts as a lower-bound as it itself does not induce any performative drift, while the TC acts like an upper-bound given that it induces performative drift to a point of saturation which is akin to the performatively stable states described in Perdomo et al. \cite{PerdomoZrnicMendler2020}. We have included a figure in the Appendix that demonstrates these effects.

In Figure \ref{fig:CLASSIFIERS}, it can be seen that, as $mix$ is increased, the detection rate of the CB-PDD decreases. This is due to fewer instances being classified in accordance with the CheckerBoard detector, creating weaker feedback loops which are harder to detect. This effect is less noticeable in the RC which does not induce PD, allowing up to $75\%$ of the instances to be given to it before seeing a significant drop in detection rate. The opposite is true for the TC where a drop-off in detection rate is noticeable at $mix=0.25$. This is due to the rate at which saturation occurs in the TC. Our results indicate that performative drift must be detected before the predictive model induces a stable state. This is because in such a stable state, PD does not actually occur as the distribution of incoming instances remains constant. 

\begin{figure}[t]
\centering
\includegraphics[width=0.85\columnwidth]{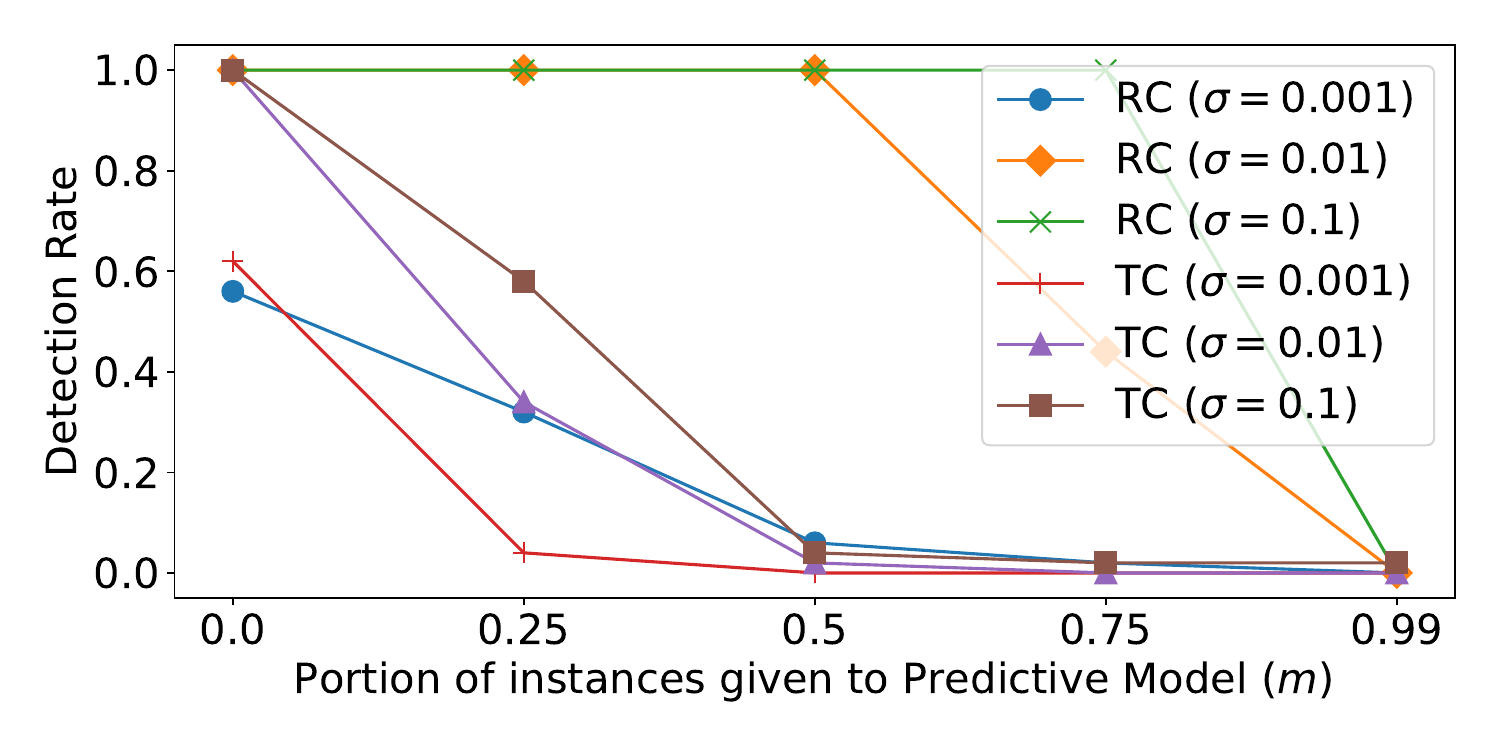}
\caption{shows the detection rate of CB-PDD when deployed in tandem with a predictive model. $m$ describes the portion of instances given to the predictive model. These results show that as $m$ increases, the detection rate decreases. The trend is more noticeable in the Threshold Classifier (TC) which induces its own PD in comparison to a Random Classifier (RC) which induces no PD.}
\label{fig:CLASSIFIERS}
\end{figure}

We have shown that while CB-PDD can be used with a predictive model, it works best in isolation. A potential solution to this problem is a dynamic mixing strategy whereby $mix$ initially starts at $0.0$ and is slowly increased to $1.0$ over $T$ instances. This allows CB-PDD to, initially, induce performative drift in a predictable manner which may make it easier to detect. Additional details regarding these experiments, including a method to reduce that rate at which saturation occurs, can be found in the Appendix.

\subsection{Intrinsic Drift}

Recall that PD is a type of Concept Drift that occurs as result of a deployed predictive model. All other types of Concept Drift are intrinsic drift. In a perfect world, performative and intrinsic drift could not coexist but, that is not the case. In order for CB-PDD to be an effective PD detector, it needs to be robust against intrinsic drift. To investigate this, we conducted experiments using two-types of intrinsic drift.

\subsubsection{Sudden Intrinsic Drift}
Sudden drift is caused by an abrupt change to the underlying data generating process. To simulate this in our data generator, we define a parameter $E$ which describes the number of drift events that will occur within a simulation run. When one of these events occurs, each centroid in the data generator is assigned a new position in the feature space $\sim U(-1,1)$.

\begin{figure}[t]
\centering
\includegraphics[width=0.85\columnwidth]{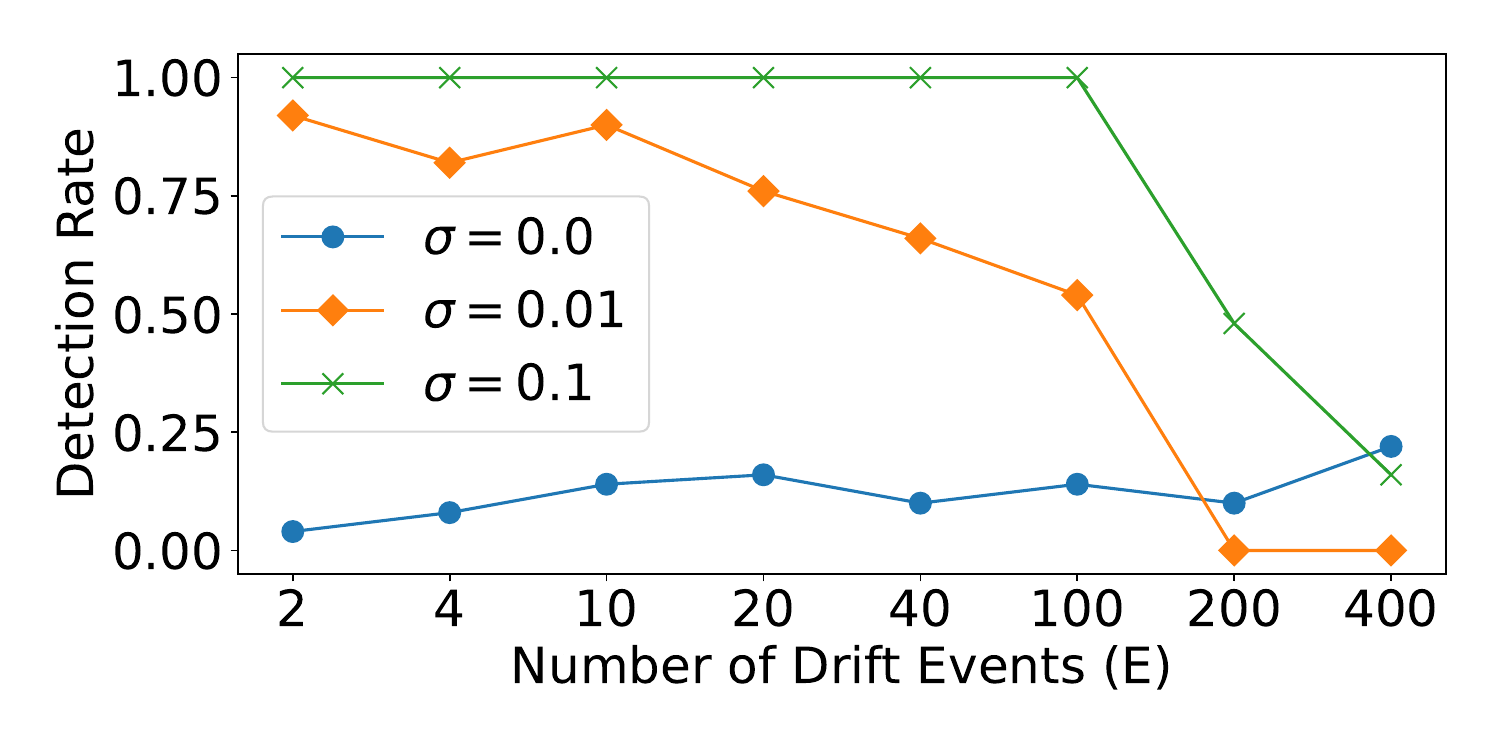}
\caption{shows the effect sudden intrinsic drift has on the detection rate of CB-PDD across various PD strengths ($\sigma$). CB-PDD appears resilient to sudden drift except in cases where the number of drift events ($E$) occur at a rate greater than the trial length parameter ($\tau = 1000$).}
\label{fig:SUDDEN}
\end{figure}

Results are shown in Figure \ref{fig:SUDDEN}. CB-PDD achieves a low detection rate when $\sigma = 0.0$, indicating that CB-PDD is robust to sudden drift. Additionally, in both high and low intensity PD settings $\sigma=\{0.01, 0.1\}$, there is manageable deterioration in the detection rate as the number of drift events increases. However, when the frequency of drift events $< \tau$, the detection rate decreases significantly. This result is predictable as it impossible for CB-PDD to accurately determine density changes when the underlying data generating process is so chaotic. From a practical perspective, results are positive. CB-PDD appears robust against sudden intrinsic drift. However, caution should be taken when choosing an appropriate $\tau$. Earlier results (Figure \ref{fig:TAUEXPLORE}) show that increasing $\tau$ increases detection rate, but this will not necessarily be the case if the rate at which sudden drift events occur in the system are more frequent than $\tau$

\subsubsection{Incremental Intrinsic Drift}

Incremental Drift occurs when the underlying data generating process changes slowly over time. To simulate this in our data generator, centroids are assigned a velocity $v_i \sim U(-1, 1)$. After an instance is generated, each centroid is moved in the feature space: $c_i = c_i + v_i * d$ where $d = 0.0001$ controls the maximum velocity of the centroids. We then utilize $E$ which describes the number of drift events that occur within a simulation run. When one of these events occurs, the velocity of each centroid is assigned a new value $\sim U(-1,1)$.

\begin{figure}[t]
\centering
\includegraphics[width=0.85\columnwidth]{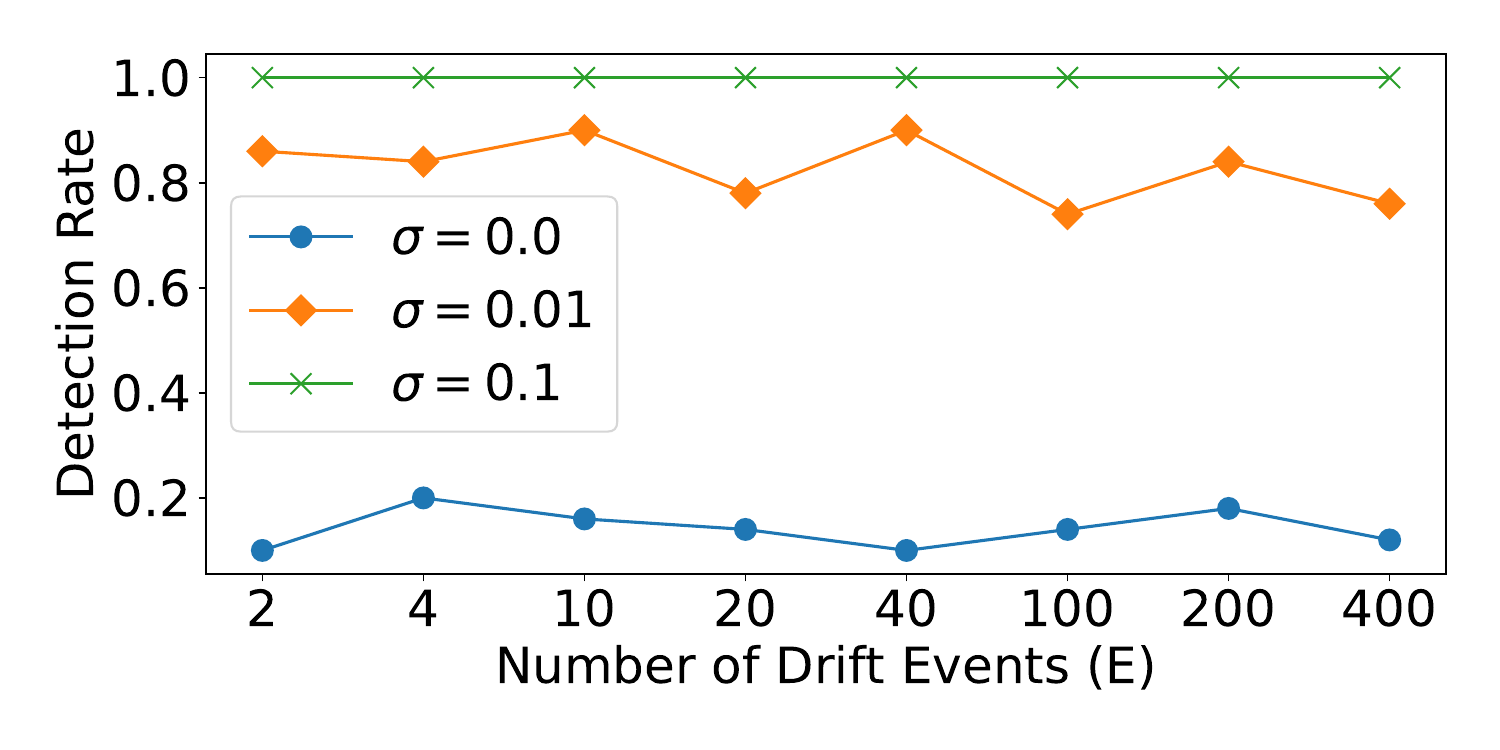}
\caption{shows the effect incremental intrinsic drift has on the detection rate of CB-PDD across various PD strengths ($\sigma$). These results suggest that CB-PDD is resilient to incremental drift even as the number of drift events ($E$) increases.}
\label{fig:GRADUAL}
\end{figure}

Shown in Figure \ref{fig:GRADUAL}, CB-PDD is robust to Intrinsic Drift in settings when there is no PD (when $\sigma = 0.0$, the false detection rate is low). Additionally, we do not observe the same degradation in detection rate as the number of drift events increases. Results suggest that CB-PDD is immune to the effects of incremental drift. This is perhaps overly optimistic, and while CB-PDD does have resilience to incremental drift, it is quite easy to construct scenarios in which that is not the case. Consider a fail case where a cluster of instances drift from one CheckerBoard group to another before the end of a trial period. In such a scenario, CB-PDD would interpret this as a density change caused by PD. This could result in a false detection event (i.e. a false positive) if no PD is present, and cause no detection event to be triggered if PD is present (i.e. a false negative). We bring attention to this issue, because it further illustrates the importance of choosing an appropriate $\tau$ which should not overlap with these intrinsic drift phenomena. A diagram of the aforementioned fail case and additional experiments with incremental drift in scenarios with higher $\epsilon$ are included in the Appendix.

\subsection{Traditional Drift Detectors}

Given that PD is a type Concept Drift, it is natural to assume that traditional Concept Drift Detectors would also work in detecting PD. We test this assumption by evaluating two popular drift detectors: ADWIN \cite{BifetGavalda2007}, a state-of-the-art featured-based detector, and DDM \cite{GamaMedasCastillo2004}, an error-based detector. We use the implementations provided by the \textit{river} framework \cite{MontielHalfordMastelini2021}.

\begin{figure}[t]
\centering
\includegraphics[width=0.85\columnwidth]{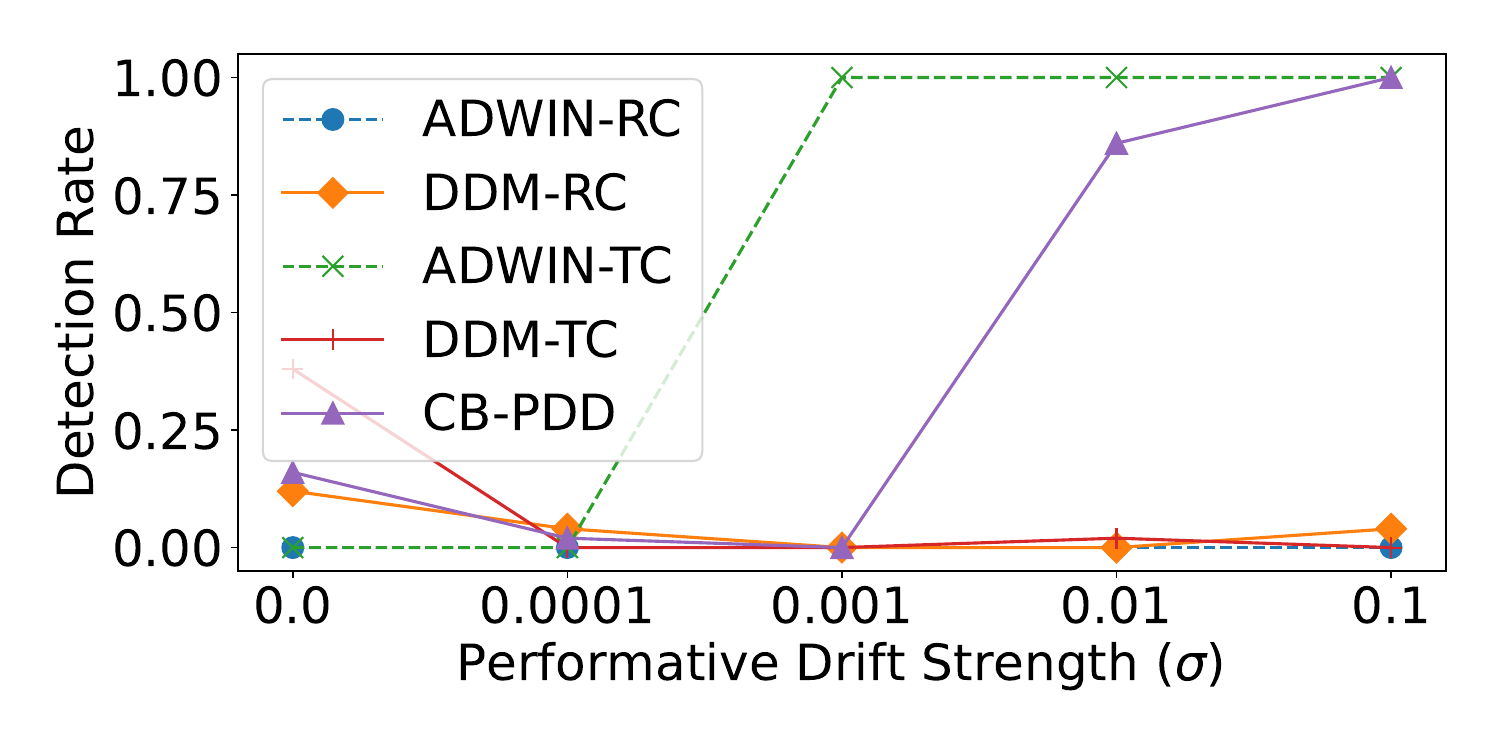}
\caption{compares performative drift strength ($\sigma$) to the detection rate of traditional Concept Drift detectors. Results show that traditional drift detectors are reliant on the predictive model they are paired with. For example, neither ADWIN or DDM can detect PD when a Random Classifier (RC) is used, because a RC induces no PD.}
\label{fig:OTHER}
\end{figure}

Unlike CB-PDD, both ADWIN and DDM must be used with a predictive model. We make use of the previously defined Random Classifer (RC) and Threshold Classifier (TC). Our results (Figure \ref{fig:OTHER}) show that neither ADWIN nor DDM are able to reliably detect PD (across any $\sigma$) when used with RC. This is because unlike CB-PDD, the other detectors don't induce their own PD, and given that the RC, doesn't induce PD either, PD is near-impossible to detect. When the TC is used, ADWIN is able to reliably detect performative at all but the lowest $\sigma = 0.0001$ outperforming CB-PDD.

DDM failed to detect PD in all scenarios investigated. DDM is an error-based drift detector. This means that a detection event is only triggered if the error of the model increases over time. In the performative setting, feedback loops don't always increase the error-rate of a predictive model, and in some cases such as these experiments, can actually decrease the error-rate. This paints an interesting picture where we can start separate the characteristics of CB-PDD from other drift detection methods. In settings with intrinsic drift and no PD, CB-PDD has shown that it will not raise false detection events. Traditional detectors will raise detection events as they are designed to detect intrinsic drift. In settings with PD and no intrinsic drift, CB-PDD can be used reliably. Feature-based detectors such as ADWIN work well too, but error-based detectors will only work if the PD causes an increase in the error-rate of the deployed model. In settings with both performative and intrinsic drift, CB-PDD has shown that it can isolate the PD and raise detection events appropriately. Other detectors can also be used in this setting, but they come with the limitation that they cannot distinguish between performative and intrinsic drift.

\subsection{Semi-Synthetic Datasets}
For our final Experiments, we wanted to evaluate CB-PDD on real-world datasets. Given that no such datasets exist, we take existing datasets and impute them with PD. While undesirable, this does have the benefit in that we are able to control the amount of PD each dataset exhibits. We take three datasets: Water Potability ($10$ features, $3276$ instances) \cite{Kadiwal2021}, Loan Status ($11$ features, $4269$ instances) \cite{Kai2023} and Credit Card ($29$ features $284 807$ instances) \cite{DalCaelenJohnson2015}. All three datasets are binary classification tasks with the Credit Card dataset being heavily imbalanced with only $0.172\%$ of the instances belonging to class $1$. To impute PD, we use our data generator. For each dataset, a centroid is created for each instance that only produces that instance when queried. The weights of all centroids are set to $1/N$ where $N$ is the number of instances in the dataset. This is done to preserve the initial distribution of each dataset. From this setup, $T=100000$ instances are sampled as in previous experiments.

\begin{figure}[t]
\centering
\includegraphics[width=0.85\columnwidth]{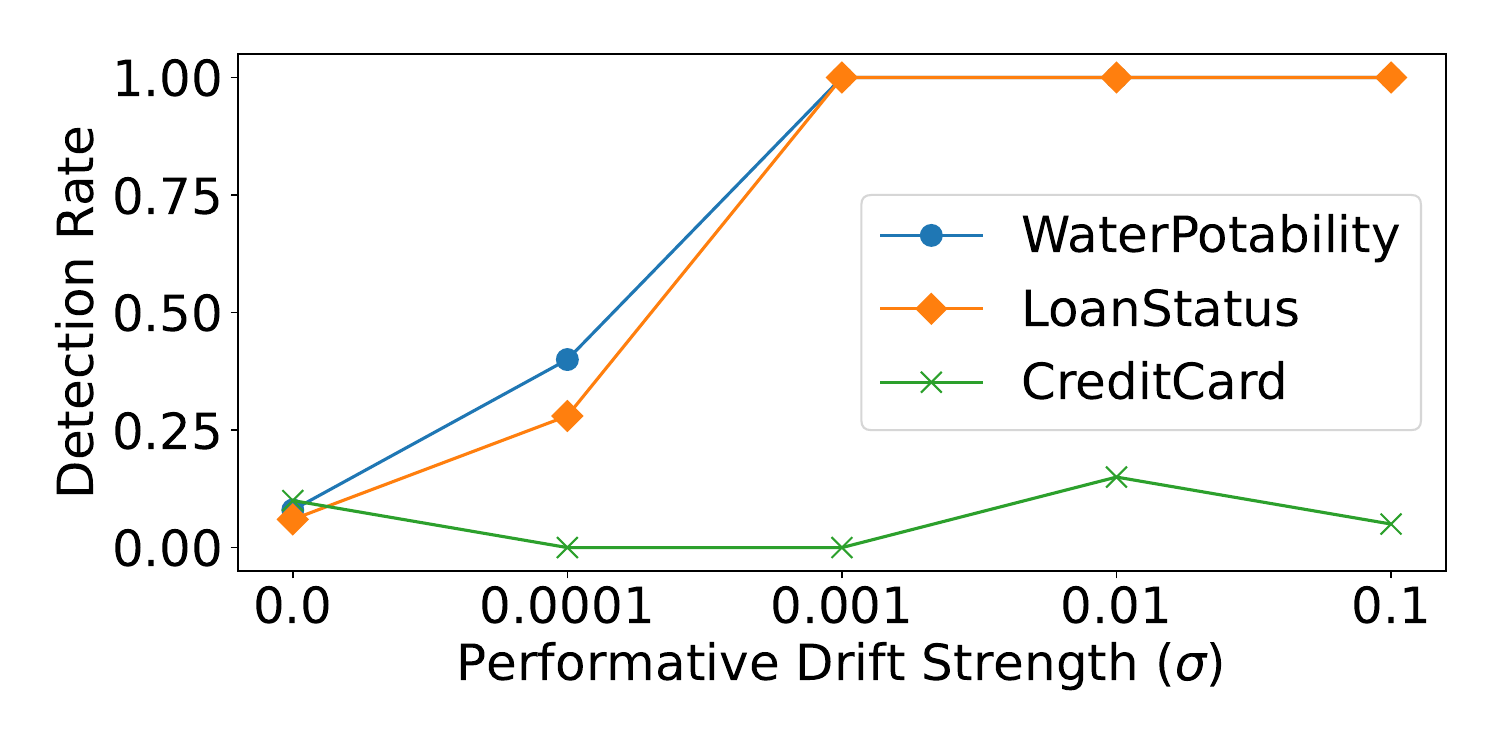}
\caption{shows the detection rate of CB-PDD on datasets with imputed PD ($\sigma$). Results show that CB-PDD can effectively detect PD in semi-synthetic datasets, but it is susceptible to class imbalances, as shown by the \textit{CreditCard} dataset.}
\label{fig:datasets}
\end{figure}

Figure \ref{fig:datasets} shows positive results. For the Water Potability and Loan Status datasets, CB-PDD achieves near-perfect detection rates for all but the smallest $\sigma = 0.0001$. These results are better than our purely-synthetic experiments. We attribute this to the consistent initial distribution that each simulation starts with. In the synthetic setting, each simulation had a random starting distribution. This allowed us to investigate a wider range of scenarios than in these experiments, but it also meant that some distributions were ill-suited for CB-PDD. These results give us confidence that CB-PDD can detect PD in a practical setting. For the Credit Card dataset, CB-PDD failed to detect drift events for both classes across all scenarios. This is due to the heavy class imbalance in the dataset indicating a limitation of our method.

\subsection{Limitations}

Experimental results are promising. CB-PDD can detect PD in a variety of settings, is resilient to intrinsic drift, and we showed why traditional concept drift detectors are insufficient tools for isolating PD. However, this paper is not without limitations. The greatest issue is the lack of real-world data. This extends to our findings that CB-PDD's trial length ($\tau$) parameter is sensitive to the intensity of PD a system experiences. We abstracted PD intensity ($\sigma$) and it not clear how the values chosen in this work translate to real-world settings. We hope this paper serves as a springboard for curating datasets with known PD which CB-PDD can be evaluated on.
Furthermore, we want to reiterate that CB-PDD is an intervention testing method. This means that in a practical setting, some portion of the incoming instances must be deliberately misclassified. If this is infeasible or unethical, another detection method would need to be developed.

\section{Related Work}

\cite{PerdomoZrnicMendler2020} pioneered Performative Prediction, but learning in the performative setting has emerged over the years under various names.  \cite{KremplHoferWebb2021,KremplBodnarHrubos2015IDA} named scenarios of prediction-induced drift as Influential Machine Learning. \cite{khritankov2023} contributed to formally modelling feedback loops that induce concept drift and other researchers such as \cite{AdamChangHaibe2020,TaoriHashimoto2023} describe performative prediction scenarios as learning with feedback loops. Interestingly, their works were less concerned about finding optimal or stable points during training, but rather on the real-world implications of performative settings, such as model trust and bias amplification. Performative Prediction has also expanded since its inception. Examples include \cite{BrownHodKlemaj2022,LiWai2022} who have done work on state-dependent performativity while \cite{NarangFaulknerDrusvyatskiy2023,PiliourasYu2023} brought performative prediction to the multi-agent setting. Lastly, the notion of performativity is also well known in other related fields such as Recommender Systems \cite{MansouryAbdollahpouriPechenizkiy2020} where ranking recommendations naturally influences the choices made by consumers, and Adversarial Learning \cite{LancewickiRosenbergMansour2022} (e.g. Spam Detection) where detectors must adapt to malicious entities who themselves are adapting to the detectors' predictions.

\section{Conclusions}

In this work, we investigated Performative Drift (PD), a type of Concept Drift that arises from feedback loops induced by predictions made by a predictive model. We defined PD in terms of both Performative Prediction \cite{PerdomoZrnicMendler2020} and traditional Concept Drift research. We then introduced \textit{CheckerBoard Performative Drift Detector (CB-PDD)}, a first-of-its-kind PD detection algorithm. We then investigated CB-PDD across a range of scenarios using a synthetic data generator. These experiments included sensitivity analysis of CB-PDD's most important parameters, an investigation of CB-PDD's capabilities when deployed with a predictive model, an analysis of CB-PDD's resilience to intrinsic drift, and a comparison of CB-PDD to traditional drift detection algorithms. Lastly, we investigated the efficacy of CB-PDD on datasets with imputed PD. 

Overall, our results are positive with CB-PDD showing high efficacy, low false detection rates, resilience to intrinsic drift, comparability to other drift detection techniques and an ability to effectively detect PD in semi-synthetic datasets. The main limitation of this work is that we only investigated CB-PDD on binary classification tasks with self-fulfilling feedback loops. Additionally, CB-PDD falters when applied to imbalanced datasets and has yet to be tested in a practical setting. Issues which we aim to address with future work.

\section*{Acknowledgements}
We would like to thank the participants of the Dagstuhl Seminar 20372  "Beyond Adaptation: Understanding Distributional Changes" held in September 2020 for the valuable discussions on the topic of drift, and specifically on influential predictions, and the types and origins of drift. In particular, we would like to thank 
Battista Biggio at University of Cagliari,  
Matthias Deliano at Leibniz Institute for Neurobiology, 
Barbara Hammer at Bielefeld University, 
Eyke H{\"u}llermeier at LMU Munich, 
Vincent Lemaire at Orange Labs, 
Michele Sebag at  Universit{\'e} Paris Saclay, 
Myra Spiliopoulou at Magdeburg University, 
Jerzy Stefanowski at Poznan University of Technology, 
Dirk Tasche at the Swiss Financial Market Supervisory Authority, 
Oskar J. Gstrein and Andrej J. Zwitter at Rijksuniversiteit Groningen, 
and Mark Nelson and Margarita Quihuis from the Peace Innovation Lab at Stanford University.

\appendix

\section{Pseudocode for CB-PDD}

See Algorithms \ref{alg:CheckerBoardPrediction} and \ref{alg:CBPDD23} for Pseudocode of CB-PDD.

\begin{algorithm}[tb]
\caption{Pseudocode for CheckerBoard Predictor (Used in Stage 1) for a Binary Classification task.}
\label{alg:CheckerBoardPrediction}
\textbf{Input}: Single feature value $x$  \\
\textbf{Parameters}: Feature split $f$, trial length $\tau$\\
\textbf{Output}: Predicted label $\hat{y}$
\begin{algorithmic}[1] 
\STATE $predictors=[1, 0, 0, 1]$.
\STATE $f_{id} = \lfloor x / f \rfloor \mod 2$
\STATE $t_{id} = \lfloor x / \tau \rfloor \mod 2$
\STATE $\hat{y} = predictors[f_{id} + 2t_{id}]$
\STATE \textbf{return} $\hat{y}$
\end{algorithmic}
\end{algorithm}

\begin{algorithm}[tb]
\caption{Stages 2 and 3 in CB-PDD. This process is repeated for each class in the classification task.}
\label{alg:CBPDD23}
\textbf{Input}: Class label $c$, set of predictions $\hat{Y}$ and labels $Y$ for instances classified by the CheckerBoard Predictor \\
\textbf{Parameters}: Total number of instances $T$, trial length $\tau$, confidence value $\alpha$ and window size $w$ \\
\textbf{Output}: True if a drift event was detected for class $c$, else False.
\begin{algorithmic}[1] 
\STATE $A=[\,]$ and $B=[\,]$.
\FOR{t in range($T / \tau$)}
\STATE start = $\tau * t$
\STATE end = $\tau * (t+1)$

\STATE $d_1 = \{ \hat{y}_i \in \hat{Y} \, \vert \verb| start| \leq i < \verb|start|+w \verb| and | \hat{y}_i = c\}$
\STATE $d_2 = \{ \hat{y}_i \in \hat{Y} \, \vert \verb| end|-w \leq i < \verb|end and | \hat{y}_i = c\}$

\STATE $corr_1 = \{ \hat{y}_i \in d_1 \, \vert \, \hat{y}_i = y_i \}$
\STATE $corr_2 = \{ \hat{y}_i \in d_2 \, \vert \, \hat{y}_i = y_i \}$

\STATE $a = (\vert corr_2 \vert / \vert d_2 \vert) - (\vert corr_1 \vert / \vert d_1 \vert)$
\STATE $b = -a$

\STATE A.append(a) and B.append(b)

\ENDFOR
\STATE $p = stats\_test(A, B).pvalue$
\STATE \textbf{return} True if $p < \alpha$ else False
\end{algorithmic}
\end{algorithm}

\section{Full Experimental Details}
\subsection{Code Availability and Compute Resources}

All code used in this work can be found at the following url: \url{https://github.com/BrandonGower-Winter/CheckerboardDetection/releases/tag/v1.0}. Instructions on how to run the code are included in the \textit{README.md} file. The computer used to run all experiments was a MacBook Pro (2 GHz Quad-Core Intel Core i5) with Intel Iris Plus Graphics (1536 MB) and 16GB of RAM

\subsection{Parameter List}

A list of all parameters used in our experiments are shown in Table \ref{tab:PARAMETERS}. For the exploration of the $\tau$ and $f$ parameters, we also experimented with $\sigma = \{0.0005, 0.005, 0.05\}$ to further understand the dynamics of said parameters. In settings with high $\epsilon$, the Gaussian Spread parameter was increased to $0.15$ and $C$ was decreased to $10$.

\begin{table}[t]
\centering
\resizebox{.5\columnwidth}{!}{
\begin{tabular}{l|l}
    \hline
    Parameter & Values \\
    \hline
    n (Repetitions) & 50 \\
    T & 100 000 \\
    C (centroids per label) & 1000 \\
    $\tau$ & 1000 \\
    $f$ & 1.0 \\
    $\sigma$ & $\{0.0, 0.0001, 0.001, 0.01, 0.1 \}$ \\
    $\alpha$ & 0.01 \\
    Statistical Test & Mann-Whitney U Test \\
    Feedback Type & Self-Fulfilling \\
    Gaussian Spread & 0.01 \\
    \hline
\end{tabular}}
\caption{Full Parameter List of Experiments conducted in this work.}
\label{tab:PARAMETERS}
\end{table}

\subsubsection{Dataset Preparation and Motivation}

For our final Experiments, we wanted to evaluate CB-PDD on real-world datasets. Given that no such datasets exist, we took existing datasets and imputed them with PD. While undesirable, this does have the benefit in that we are able to control the amount of PD each dataset exhibits. We took three datasets: Water Potability ($10$ features, $3276$ instances) \cite{Kadiwal2021}, Loan Status ($11$ features, $4269$ instances) \cite{Kai2023} and Credit Card ($29$ features $284 807$ instances) \cite{DalCaelenJohnson2015}. To prepare each dataset, all categorical features were removed, all numerical values were normalized between $[-1.0, 1.0]$ if negative values were permitted. All other numerical values were normalized between [0.0, 1.0] if negative values were not permitted.

Each dataset was selected because, in the real-world, it would have the potential to exhibit performative drift. In a regulatory environment, water standards might not be monitored effectively, and if not detected, could cause a feedback loop whereby more entities try to skirt such regulations. In a loan determination setting, successful loan application may cause more entities (with similar properties) to apply for similar loans known that they will get them. In the fraud detection setting, not detecting said fraud can cause an uptick in fraud cases.

\subsection{Visualization of Initial Distributions}

There is no consensus on how to model PD within a data-stream and, to the best of our knowledge, there exists no datasets with detected PD or data generators that support PD. In \cite{PerdomoZrnicMendler2020}, they use the deployed model's parameters $\theta$ to model performative drift over time. This approach is limited in that it assumes that the user has access to $\theta$, and is not compatible with approaches that impose alternative classification regimes such as CB-PDD.

To address this, we developed a data generator for performative settings. Inspired by \textit{Random Radial Basis Function (RandomRBF)} generators \cite{MontielReadBifet2018}, our generator is initialized by specifying $C$ centroids within a feature space. Each centroid $c_i$ is given a Gaussian distribution, a label $y_i$ and a weight $w_i$. When an instance is requested, roulette wheel selection is performed \cite{LipowskiLipowska2012} using the weights. The Gaussian of the selected centroid is then sampled generating an instance $x$ which is assigned the corresponding label $y_i$.

\begin{figure}[t]
\centering
\includegraphics[width=0.9\columnwidth]{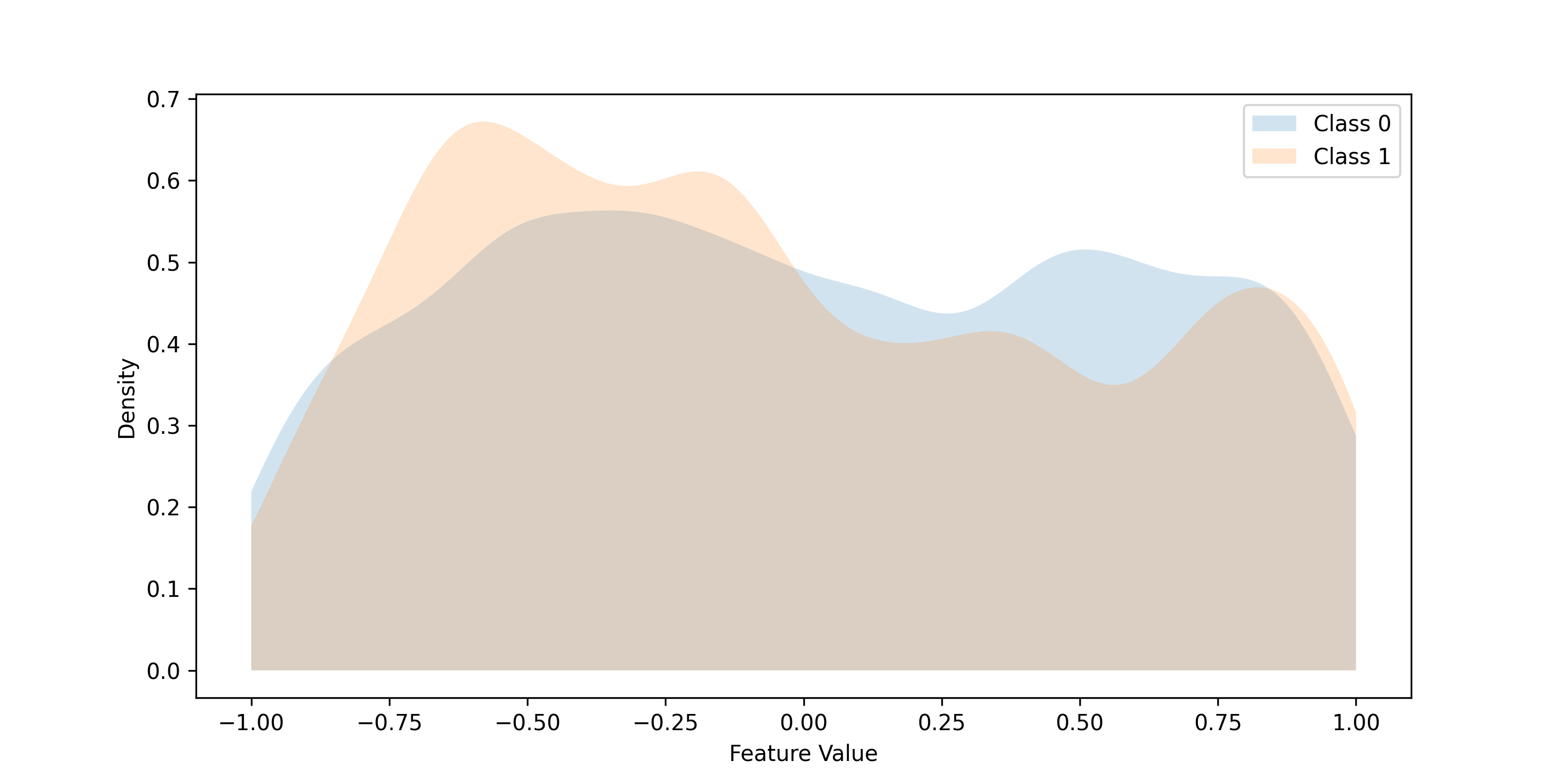}
\caption{Example Initial Distribution One.}
\label{fig:DIST1}
\end{figure}

\begin{figure}[t]
\centering
\includegraphics[width=0.9\columnwidth]{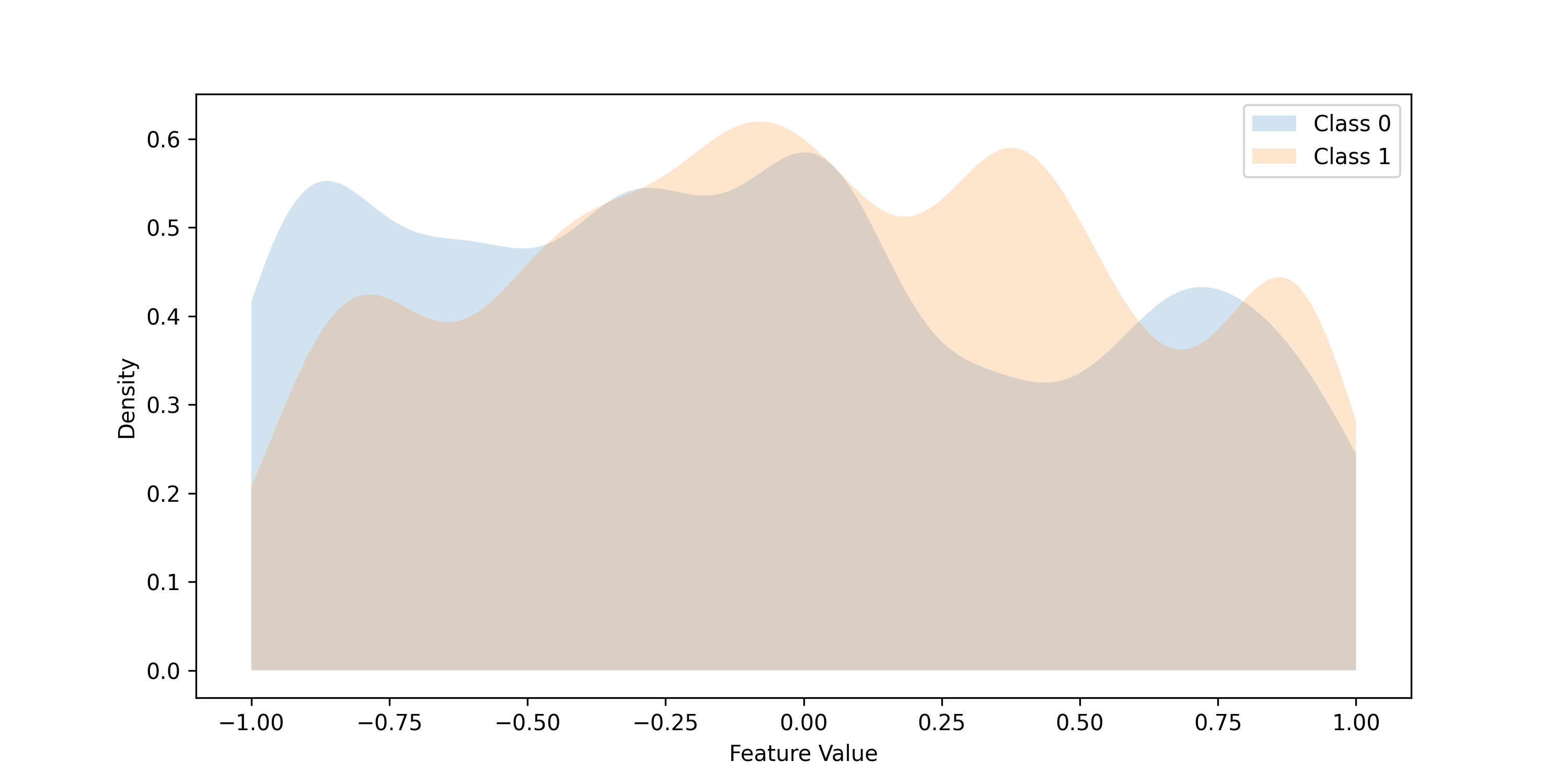}
\caption{Example Initial Distribution Two.}
\label{fig:DIST2}
\end{figure}

\begin{figure}[t]
\centering
\includegraphics[width=0.9\columnwidth]{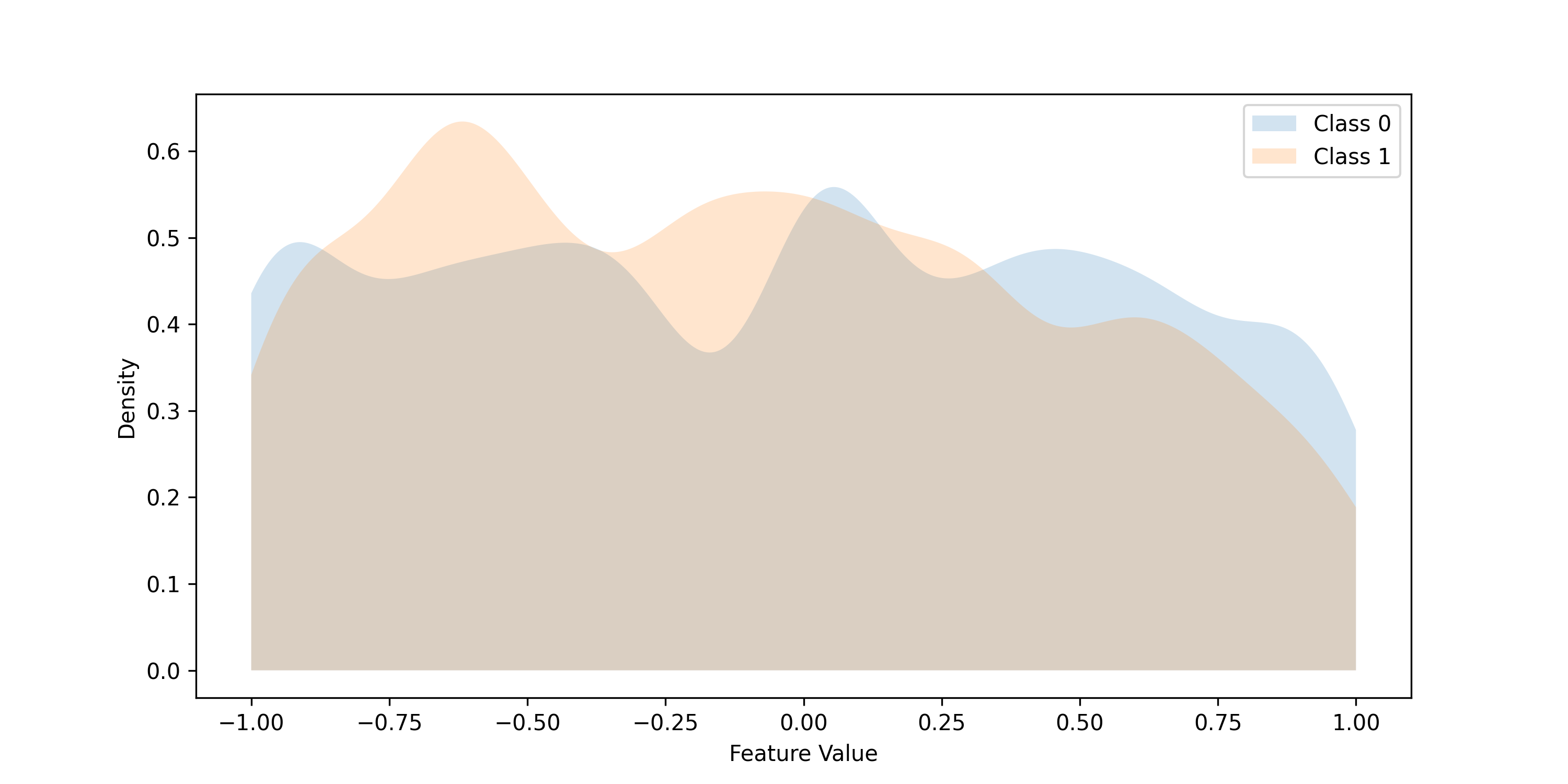}
\caption{Example Initial Distribution Three.}
\label{fig:DIST3}
\end{figure}

\begin{figure}[t]
\centering
\includegraphics[width=0.9\columnwidth]{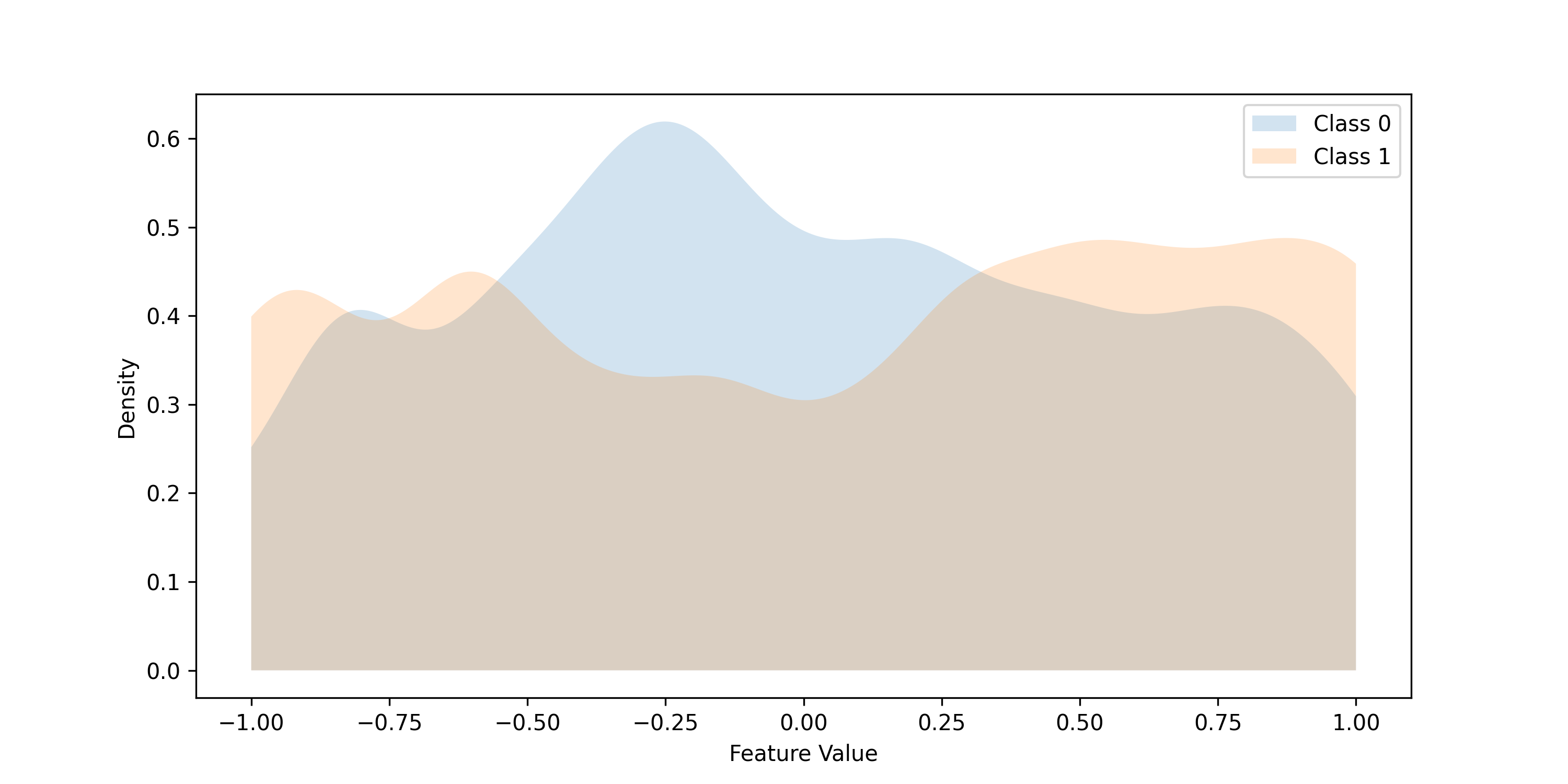}
\caption{Example Initial Distribution Four.}
\label{fig:DIST4}
\end{figure}

In order to simulate PD, a user is required to supply a predictor to the data generator as well as the intended feedback loop behaviour for each class (i.e. The target prediction $\bar{y}$ for each class and whether a self-defeating or self-fulfilling feedback loop should be used). When an instance is generated, the predictor is queried and its prediction $\hat{y}$ is stored. If $\hat{y} = \bar{y}$, the weight $w_i$ of centroid $c_i$ is modified. In the case of a self-fulling feedback loop, $w_i = w_i + \sigma$. In the case of a self-defeating feedback loop, $w_i = w_i - \sigma$ where $\sigma$ is a user-defined parameter that determines the performative drift strength of the system. Using this approach, we are able to create a variety of initial conditions, and performative drift settings which we use to evaluate CB-PDD. Figures \ref{fig:DIST1}, \ref{fig:DIST2}, \ref{fig:DIST3} and \ref{fig:DIST4} showcase examples of these initial distributions.

\section{Additional Results}

\subsection{Self-Defeating Feedback Loops}

In this work, we primarily focused on Self-Fulling feedback loops. As discussed in the main paper, these are not the only type of feedback loops. The other main type is the Self-Defeating loop whereby induced feedback loops cause a decrease in instance density (as opposed to an increase in self-fulfilling settings). To demonstrate that CB-PDD does work in self-defeating settings, we conducted experiments which are summarized in Table \ref{tab:DEFEATING}. All parameters detailed in Table \ref{tab:PARAMETERS} were used in these experiments, except that $10$ centroids were used with a Gaussian Spread of $0.15$.

\begin{table}[t]
\centering
\resizebox{.5\columnwidth}{!}{
\begin{tabular}{l|l}
    \hline
    $\sigma$ & Detection Rate \\
    \hline
    0.0 & 0.1 \\
    0.0001 & 0.04 \\
    0.001 & 0.5 \\
    0.01 & 1.0 \\
    0.1 &  0.94 \\
    \hline
\end{tabular}}
\caption{Detection Rate of CB-PDD in our Self-Defeating Experiments. In general, the same trend is observed here as in the self-sulling feedback setting. Overall, CB-PDD shows high detection rate in settings were the performative drift strength is strong.}
\label{tab:DEFEATING}
\end{table}

In general, our results show similar trends to our self-fulfilling experiments. As performative drift strength ($\sigma$) increases, so does the detection rate. A low false detection rate ($\sigma = 0.0$) is also observed. These results are promising as they suggest that CB-PDD is likely to be as effective in self-defeating feedback loops settings, although future work will expand upon these results to investigate edge-cases and find the limits of CB-PDD

\subsection{Case Where Only One Class is Performative}

In this work, all of our experiments assumed that both classes in the detection task exhibited performative drift. It is conceivable that scenarios where only one or some classes exhibit performative drift exist.  We conducted additional experiments where only Class 1 is performative (not Class 0). The results of which are summarized in Table \ref{tab:ONECLASS}. All parameters detailed in Table \ref{tab:PARAMETERS} were used in these experiments, except that $10$ centroids were used with a Gaussian Spread of $0.15$.

\begin{table}[t]
\centering
\resizebox{.5\columnwidth}{!}{
\begin{tabular}{l|l}
    \hline
    $\sigma$ & Detection Rate \\
    \hline
    0.0 & 0.1 \\
    0.0001 & 0.1 \\
    0.001 & 0.72 \\
    0.01 & 0.94 \\
    0.1 &  0.96 \\
    \hline
\end{tabular}}
\caption{Detection Rate of CB-PDD when only Class 1 is performative. The results show similar trends to that of our main results. As performative drift strength increases ($\sigma$), so does the detection rate}
\label{tab:ONECLASS}
\end{table}

In general, our results show similar trends to our main experiments. As performative drift strength ($\sigma$) increases, so does the detection rate. A low false detection rate ($\sigma = 0.0$) is also observed. Interestingly, there were cases, as $\sigma$ increases, where Class 0 was falsely detected. It is not clear what the ramifications are from such a false detection, given that the deployed predictive model will still need to account for performativity in Class 1. Additionally, these results also suggest that it is easier to detect only one performative class in settings with lower $\sigma$. Intuitively this makes sense, if only one class is performative, density changes are easier to detect. However, further experimentation is need to ascertain if this rings true across a wider range of scenarios, and more importantly, in a practical setting.

\subsection{Visualization of Threshold and Random Classifier Drift}

\begin{figure}[t]
\centering
\includegraphics[width=0.9\columnwidth]{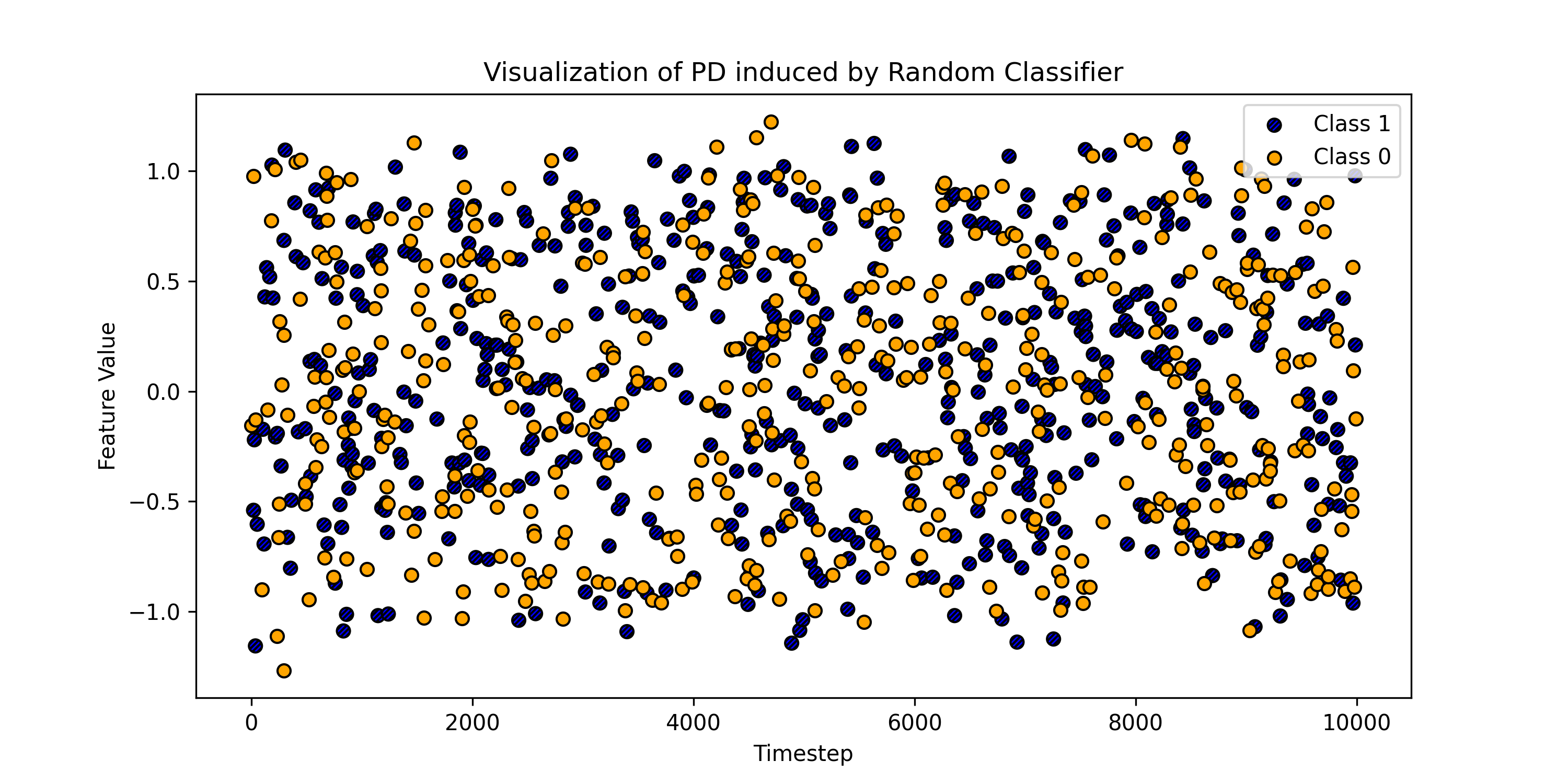}
\caption{Example of a data-stream (with Performative Drift) when a Random Classifier is used. The figure shows no performative effect because a Random Classifier does not induce any feedback loops, and thus no Performative Drift.}
\label{fig:RCViz}
\end{figure}

\begin{figure}[t]
\centering
\includegraphics[width=0.9\columnwidth]{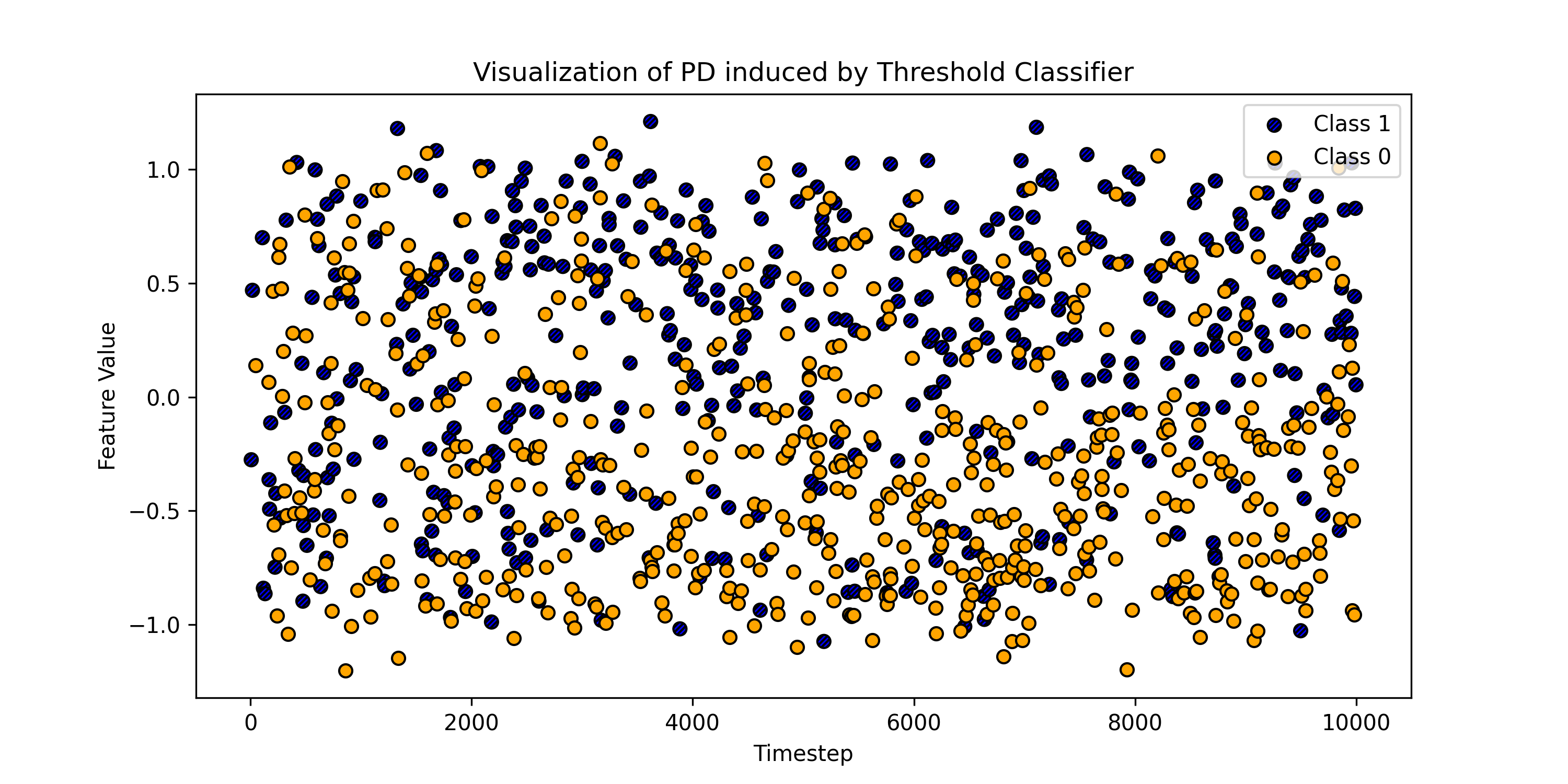}
\caption{Example of a data-stream (with Performative Drift) when a Threshold Classifier is used. The figure shows a saturation effect whereby the instances with $x > 0$ are predominately of Class 1 while instances with $x < 0$ are mostly Class 0. This is because a Threshold Classifier induces a feedback loop through the persistent classification of instances.}
\label{fig:TCViz}
\end{figure}

We investigated two types of predictive models: The first is a simple Random Classifier (RC) which randomly predicts which class ($1$ or $0$) an incoming instance belongs too. The second model is a Threshold Classifier (TC) which predicts an instance's label in accordance with the following decision rule: $1 \verb| if | x > 0  \verb| else | 0$. The RC acts as a lower-bound as it itself does not induce any performative drift, while the TC acts like an upper-bound given that it induces performative drift to a point of saturation which is akin to the performatively stable states described in \cite{PerdomoZrnicMendler2020}. Figures \ref{fig:RCViz} and \ref{fig:TCViz} demonstrate these effects.

\subsection{Incremental Drift with High $\epsilon$}

\begin{figure}[t]
\centering
\includegraphics[width=0.9\columnwidth]{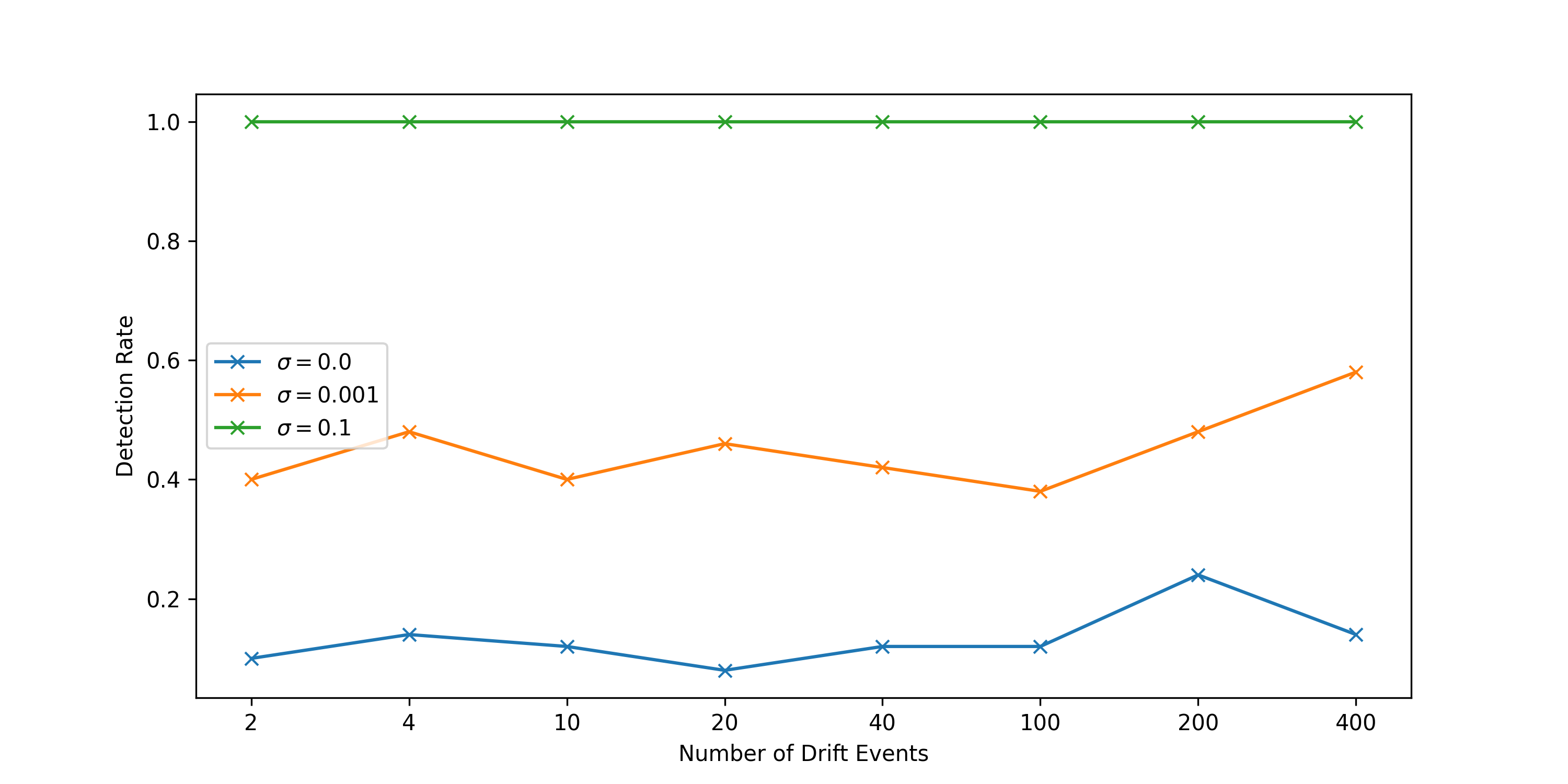}
\caption{shows the effect incremental intrinsic drift has on the detection rate of CB-PDD across various PD strengths ($\sigma$). These results suggest that CB-PDD is resilient to incremental drift even as the number of drift events ($E$) increases.}
\label{fig:HIGHEPS}
\end{figure}

In our main paper, we investigated how incremental drift affected the detection rate of CB-PDD. Overall, our findings suggested that CB-PDD, was resistant to incremental drift. We expanded on these findings by running additional experiments with high $\epsilon$. In these settings, feedback loops induced by instance $x$ might cause instances that do not look exactly like $x$ to appear. The higher $\epsilon$, the greater this dissimilarity can appear. For these experiments, we use the parameters in Table \ref{tab:PARAMETERS}, but $C$ was reduced to $10$ and Gaussian Spread to $0.15$ to simulate a higher $\epsilon$ setting. Figure \ref{fig:HIGHEPS} reports on our findings.

Overall, these results show both a similar and dissimilar trend to our original experiments. As in our main experiments, the number of incremental drift seems to have little effect on CB-PDD. However, the higher $\epsilon$ setting has caused the detection rate of the $\sigma = 0.01$ setting to decrease. This is due to range at which future instances can be produced at. When an instance near a border region between two classification groups induces a feedback loop, the higher, a future instance may fall into a different group, thus reducing the effectiveness of the feedback loop, making it harder for CB-PDD to detect. This is exasperated in drift settings where the moving clusters of instances can cause group overlaps. This effect is highlighted in more detail in the next section. The implications from these findings just reinforce our main findings where CB-PDD is an effected performative drift detector, but it is heavily reliant on the correct use of the $f$ and $\tau$ parameters to minimize the effects of intrinsic drift and the dissimiliarity at which future feedback-induced instances can be generated at ($\epsilon$).

\subsection{Fail Case for Intrinsic Drift}

\begin{figure}[t]
\centering
\includegraphics[width=0.9\columnwidth]{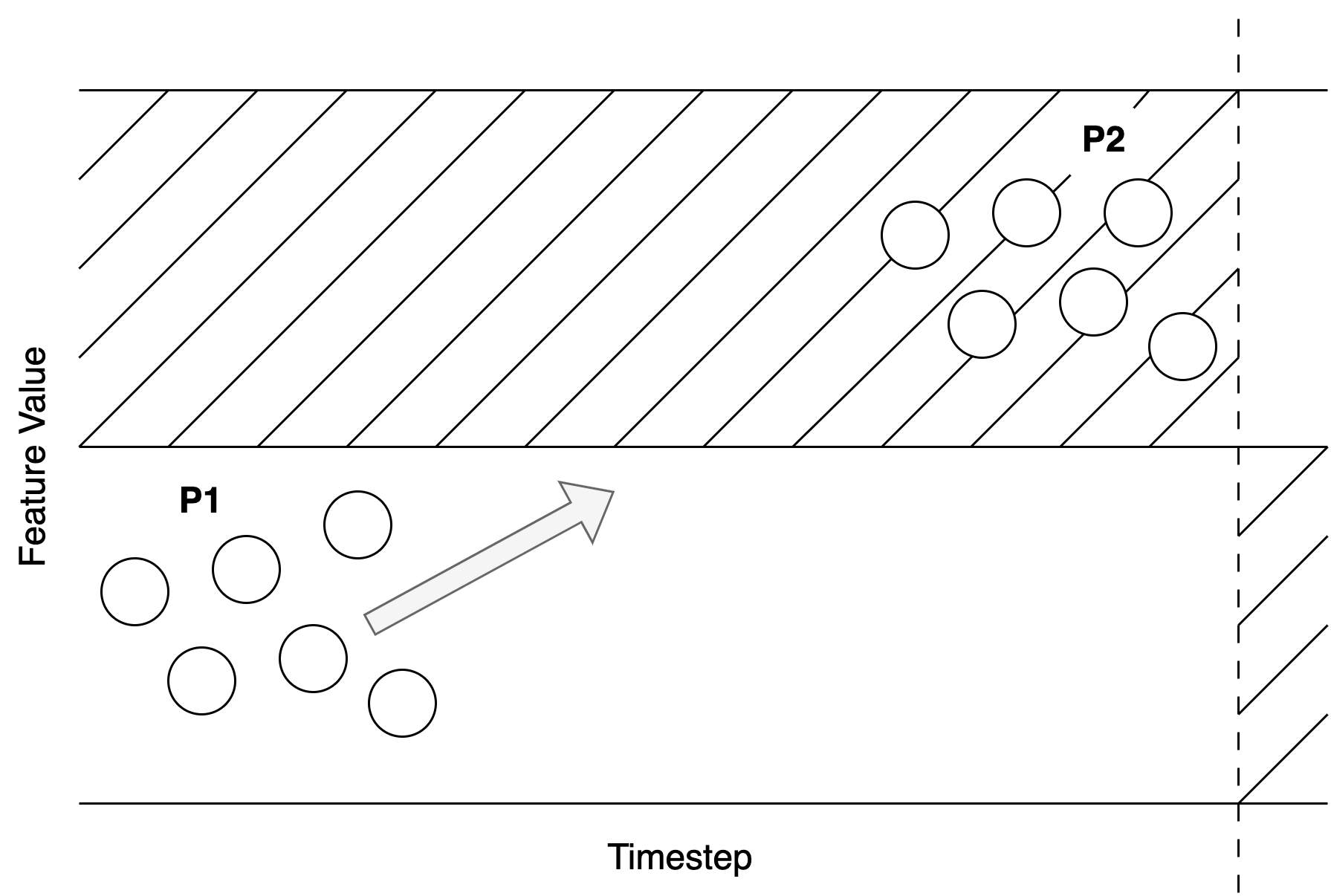}
\caption{Example of CB-PDD fail case in setting with incremental intrinsic drift. In this example, a cluster of instances drift from P1 to P2 before the end of a trial period. This will result in a density change which CB-PDD can mistake for a drift event in the case of no performative drift, or no drift event in the case where performative drift is present.}
\label{fig:IDRIFTFAILCASE}
\end{figure}

As shown in our main paper, CB-PDD is robust to Intrinsic Drift in settings when there is no performative drift (when $\sigma = 0.0$, the false detection rate is low). Additionally, we do not observe the same degradation in detection rate as the number of drift events increases. Our results suggest that CB-PDD is immune to the effects of incremental drift. This is perhaps overly optimistic, and while it is true that CB-PDD does have resilience to incremental drift, it is quite easy to construct scenarios in which that is not the case. Consider a fail case where a cluster of instances drift from one CheckerBoard group to another before the end of a trial period. In such a scenario, CB-PDD would interpret this as a density change caused by PD. This could result in a false detection event (i.e. a false positive) if no PD is present, and cause no detection event to be triggered if PD is present (i.e. a false negative). We bring attention to this issue, because it further illustrates the importance of choosing an appropriate $\tau$ which should not overlap with these intrinsic drift phenomena. A diagram of the aforementioned fail case is shown in Figure \ref{fig:IDRIFTFAILCASE}.

\subsection{Reducing the Saturation Effect of Classifiers}

In our main paper, we found (in Section "CB-PDD with Classifiers", that CB-PDD is only effective in detecting performative drift before a deployed classifier induces a stable-state. One way to delay this effect is a bit nuanced. During development of CB-PDD, we found that ensuring that a portion of the classifiers instances are always predicted incorrectly delays the saturation effect. This is because this process reduces the intensity of the feedback loop produced by the deployed predictor. In our case, the Threshold Classifier always predicts Class 1 if instance $x > 0$. By ensuring that some portion of instances $x > 0$ are predicted as Class 0, the intensity of the feedback loop is reduced. With CB-PDD, we could do this by using an $f=0.5$. This did result in a better detection rates overall as shown in Table \ref{tab:MIX} when enough instances were given to CB-PDD ($m \leq 0.5$), but it was not able to reduce the main limitation that CB-PDD requires a substantial number of instances be dedicated to it (as opposed to the predictive model) when the predictive model induces its own performative drift.

\begin{table}[t]
\centering
\resizebox{.5\columnwidth}{!}{
\begin{tabular}{l|l|l}
    \hline
    $m$ & $f=1.0$ & $f=0.5$ \\
    \hline
    0.0 & 1.0 & 1.0 \\
    0.25 & 0.58 & 1.0 \\
    0.5 & 0.04 & 0.12 \\
    0.71 & 0.02 & 0.0 \\
    0.99 &  0.02 & 0.0 \\
    \hline
\end{tabular}}
\caption{Detection Rate of CB-PDD when paired with a Threshold Classifier. The $m$ parameter describes the probability of an instance being given to the Threshold Classifer. These results show that ensuring that some portion of the Threshold Classifiers predictions are incorrect ($f=0.5$), we are able to delay the saturation effect and increase the detection rate of CB-PDD.}
\label{tab:MIX}
\end{table}

\bibliographystyle{unsrt}  
\bibliography{main}

\end{document}